\documentclass[10pt,twocolumn,letterpaper]{article}

\usepackage{cvpr}
\usepackage{times}
\usepackage{epsfig}
\usepackage{graphicx}
\usepackage{amsmath}
\usepackage{amssymb}
\usepackage{bm}
\usepackage{multirow}
\usepackage{subfig}
\usepackage{amsthm}
\usepackage{graphicx}
\usepackage{amsfonts}
\usepackage{booktabs}
\usepackage{enumitem}
\usepackage{balance}
\usepackage{xcolor}
\graphicspath{{./imgs/}}
\newtheorem{theorem}{Theorem}

\DeclareMathOperator{\bbP}{\mathbb{P}}
\DeclareMathOperator{\bbQ}{\mathbb{Q}}
\DeclareMathOperator{\bbR}{\mathbb{R}}
\DeclareMathOperator{\bbE}{\mathbb{E}}
\DeclareMathOperator{\calX}{\mathcal{X}}

\DeclareMathOperator{\calZ}{\mathcal{Z}}
\DeclareMathOperator{\calN}{\mathcal{N}}

\usepackage[pagebackref=true,breaklinks=true,letterpaper=true,colorlinks,bookmarks=false]{hyperref}

 \cvprfinalcopy % *** Uncomment this line for the final submission

\def\cvprPaperID{6863} % *** Enter the CVPR Paper ID here

\ifcvprfinal\fi
\begin{document}

%%%%%%%%% TITLE
\title{A Characteristic Function Approach to Deep Implicit Generative Modeling}

\author{Abdul Fatir Ansari$^\dagger$, Jonathan Scarlett$^{\dagger\ddagger}$, and Harold Soh$^\dagger$\\
	$\dagger$Department of Computer Science\\
	$\ddagger$Department of Mathematics\\
National University of Singapore\\
{\tt\small \{afatir, scarlett, harold\}@comp.nus.edu.sg}
}

\maketitle
\thispagestyle{empty}

\begin{abstract}
%Various distances between probability measures have been proposed for replacing the binary classifier in the discriminator of generative adversarial networks (GANs). 
Implicit Generative Models (IGMs) such as GANs have emerged as effective data-driven models for generating samples, particularly images. In this paper, we formulate the problem of learning an IGM as minimizing the expected distance between characteristic functions. Specifically, we minimize the distance between characteristic functions of the real and generated data distributions under a suitably-chosen weighting distribution. This distance metric, which we term as the characteristic function distance (CFD), can be (approximately) computed with linear time-complexity in the number of samples, in contrast with the quadratic-time Maximum Mean Discrepancy (MMD). By replacing the discrepancy measure in the critic of a GAN with the CFD, we obtain a model that is simple to implement and stable to train. The proposed metric enjoys desirable theoretical properties including continuity and differentiability with respect to generator parameters, and continuity in the weak topology. We further propose a variation of the CFD in which the weighting distribution parameters are also optimized during training; this obviates the need for manual tuning, and leads to an improvement in test power relative to CFD. We demonstrate experimentally that our proposed method outperforms WGAN and MMD-GAN variants on a variety of unsupervised image generation benchmarks.
\end{abstract}

%\vspace*{-2.5ex}
\section{Introduction}
%\vspace*{-1ex}

% motivation
Implicit Generative Models (IGMs), such as Generative Adversarial Networks (GANs)~\cite{Goodfellow2014GenerativeAN}, seek to learn a model $\bbQ_\theta$ of an underlying data distribution $\bbP$ using samples from $\bbP$. Unlike prescribed probabilistic models, IGMs do not require a likelihood function, and thus are appealing when the data likelihood is unknown or intractable. Empirically, GANs have excelled at numerous tasks, from unsupervised image generation~\cite{karras2018style} to policy learning~\cite{ho2016generative}. 

The original GAN suffers from optimization instability and mode collapse, and often requires various ad-hoc tricks to stabilize training~\cite{radford2015unsupervised}. Subsequent research has revealed that the generator-discriminator setup in the GAN minimizes the Jensen-Shannon divergence between the real and generated data distributions; this divergence possesses discontinuities that results in uninformative gradients as $\bbQ_\theta$ approaches $\bbP$, which hampers training. Various works have since established desirable properties for a divergence that can ease GAN training, and proposed alternative training schemes~\cite{Arjovsky2017TowardsPM,salimans2016improved,Arjovsky2017WassersteinGA}, primarily using distances belonging to the Integral Probability Metric (IPM) family~\cite{muller1997integral}. One popular IPM is the kernel-based metric Maximum Mean Discrepancy (MMD), and a significant portion of recent work has focussed on deriving better MMD-GAN variants~\cite{Li2017MMDGT,Binkowski2018DemystifyingMG,arbel2018gradient,Li2019ImplicitKL}.

In this paper, we undertake a different, more elementary approach, and formulate the problem of learning an IGM as minimizing the expected distance between \emph{characteristic functions} of real and generated data distributions. Characteristic functions are widespread in probability theory and have been used for two-sample testing~\cite{heathcote1972test,epps1986omnibus,chwialkowski2015fast}, yet surprisingly, have not yet been investigated for GAN training. We find that this approach leads to a \emph{simple} and \emph{computationally-efficient} loss: the characteristic function distance (CFD). %, which is the expected value of the squared absolute difference between the characteristic functions under a weighting distribution. 
Computing CFD requires linear time in the number of samples (unlike the quadratic-time MMD), and our experimental results indicate that CFD minimization results in effective training. %Although similar measures have previously been studied for two-sample tests \cite{heathcote1972test,epps1986omnibus,chwialkowski2015fast}, %and in relation to kernel methods \cite{rahimi2008random,zhao2015fastmmd}, 
%its applicability towards training generative models has remained unexplored. 

This work provides both theoretical and empirical support for using CFD to train IGMs. We first establish that the CFD is continuous and differentiable almost everywhere with respect to the parameters of the generator, and that it satisfies continuity in the weak topology -- key properties that make it a suitable GAN metric \cite{Arjovsky2017WassersteinGA,Li2017MMDGT}. We provide novel direct proofs that supplement the existing theory on GAN training metrics. 
Algorithmically, our key idea is simple: train GANs using empirical estimates of the CFD under optimized weighting distributions. We report on systematic experiments using synthetic distributions and four benchmark image datasets (MNIST, CIFAR10, STL10, CelebA). 
Our experiments demonstrate that the CFD-based approach outperforms WGAN and MMD-GAN variants on quantitative evaluation metrics. From a practical perspective, we find the CFD-based GANs are simple to implement and stable to train. 
  %; this last finding may be somewhat surprising since the empirical CFD estimates can yield high variance gradients. Nevertheless, experiments show the CFD-based approach to outperform state-of-the art models (WGAN and MMD-GAN variants) and is competitive with WGAN-GP on quantitative evaluation metrics.
In summary, the key contributions of this work are:
\begin{itemize}[noitemsep,nolistsep]
	\item a novel approach to train implicit generative models using a loss derived from characteristic functions;
	\item theoretical results showing that the proposed loss metric is continuous and differentiable in the parameters of the generator, and satisfies continuity in the weak topology;
	\item experimental results showing that our approach leads to effective generative models favorable against state-of-the-art WGAN and MMD-GAN variants on a variety of synthetic and real-world datasets.
\end{itemize}

%\vspace*{-1ex}
\section{Probability Distances and GANs}
%\vspace*{-1ex}

%\hs{the sections are quite disconnected. Can you clarify the logical flow of your arguments? A few sentences in the intro can help set things up, e.g.: \emph{In this work, we propose novel GAN models that are trained by minimizing distances between probability distributions. We begin by reviewing the standard GAN framework and modern methods for training GANs, particularly the MMD approach which is related to our contribution.}}

%In this work, we propose novel GAN models that are trained by minimizing distances between characteristic functions of probability distributions. 

We begin by providing a brief review of the Generative Adversarial Network (GAN) framework and recent distance-based methods for training GANs. A GAN is a generative model that implicitly seeks to learn the data distribution $\mathbb{P}_{\mathcal{X}}$ given samples $\{\mathbf{x}\}_{i=1}^n$ from $\mathbb{P}_{\mathcal{X}}$. The GAN consists of a generator network $g_\theta$ and a critic network $f_\phi$ (also called the discriminator). The generator $g_\theta: \calZ \rightarrow \calX$ transforms a latent vector $\mathbf{z} \in \calZ$ sampled from a simple distribution (e.g., Gaussian) to a vector $\hat{\mathbf{x}}$ in the data space. The original GAN~\cite{Goodfellow2014GenerativeAN} was defined via an adversarial two-player game between the critic and the generator; the critic attempts to distinguish the true data samples from ones obtained from the generator, and the generator attempts to make its samples indistinguishable from the true data.  

In more recent work, this two-player game is cast as minimizing a  divergence between the real data distribution and the generated distribution. The critic $f_\phi$ evaluates some probability divergence between the true and generated samples, and is optimized to maximize this divergence. In the original GAN, the associated (implicit) distance measure is the Jensen-Shannon divergence, but alternative divergences have since been introduced, e.g., the 1-Wasserstein distance \cite{Arjovsky2017WassersteinGA,Gulrajani2017ImprovedTO}, Cramer distance \cite{Bellemare2017TheCD}, maximum mean discrepancy (MMD)  \cite{Li2017MMDGT,Binkowski2018DemystifyingMG,arbel2018gradient}, and Sobolev IPM \cite{Mroueh2017SobolevG}. Many distances proposed in the literature can be reduced to the Integral Probability Metric (IPM) framework with different restrictions on the function class. 

%\vspace*{-1ex}
\section{Characteristic Function Distance}
%\vspace*{-1ex}

In this work, we propose to train GANs using a distance metric based on characteristic functions (CFs). Letting $\bbP$ be the probability measure associated with a real-valued random variable $X$, the characteristic function $\varphi_{\bbP}: \mathbb{R}^d \rightarrow \mathbb{C}$ of $X$ is given by
\begin{align}
\varphi_{\bbP}(\mathbf{t}) = \mathbb{E}_{\mathbf{x}\sim\bbP}[e^{i\langle \mathbf{t}, \mathbf{x}\rangle}] = \int_{\mathbb{R}}e^{i\langle \mathbf{t}, \mathbf{x}\rangle}d\bbP,
\end{align}
where $\mathbf{t} \in \mathbb{R}^d$ is the input argument, and $i=\sqrt{-1}$. Characteristic functions are widespread in probability theory, and are often used as an alternative to probability density functions.  The characteristic function of a random variable completely defines it, i.e., for two distributions $\bbP$ and $\bbQ$, $\bbP = \bbQ$ if and only if $\varphi_{\bbP} = \varphi_{\bbQ}$. Unlike the density function, the characteristic function always exists, and is uniformly continuous and bounded: $|\varphi_{\bbP}(t)| \leq 1$.  
%% properties of CFs
% introduce  if we need the  bottom part. 
%If the random variable $X$ has a probability density function $f(\mathbf{x})$, then the characteristic function is the Fourier transform of $f(\mathbf{x})$. The $k$-th order moments of a distribution are given by $i^k\varphi_{\bbP}^{(k)}(0)$. 

The squared Characteristic Function Distance (CFD)~ \cite{chwialkowski2015fast,heathcote1977integrated} between two distributions $\bbP$ and $\bbQ$ is given by the weighted integrated squared error between their characteristic functions
\begin{align}
{\mathrm{CFD}_{\omega}^2(\mathbb{P},\mathbb{Q})} &= \int_{\mathbb{R}^d}\left|\varphi_\mathbb{P}(\mathbf{t}) - \varphi_\mathbb{Q}(\mathbf{t})\right|^2 \omega(\mathbf{t};\eta)d\mathbf{t},
\label{eq:cfdint}
\end{align}
where $\omega(\mathbf{t};\eta)$ is a weighting function, which we henceforth assume to be parametrized by $\eta$ and chosen such that the integral in Eq.~\eqref{eq:cfdint} converges.  When $\omega(\mathbf{t};\eta)$ is the probability density function of a distribution on $\mathbb{R}^d$, the integral in Eq. (\ref{eq:cfdint}) can be written as an expectation:
\begin{align}
{\mathrm{CFD}_{\omega}^2(\mathbb{P},\mathbb{Q})} = \mathbb{E}_{\mathbf{t}\sim\omega(\mathbf{t}; \eta)}\left[\left|\varphi_\mathbb{P}(\mathbf{t}) - \varphi_\mathbb{Q}(\mathbf{t})\right|^2\right].
\label{eq:cfdexpectation}
\end{align}

\begin{figure*}
\centering
\subfloat{\includegraphics[width=0.5\linewidth]{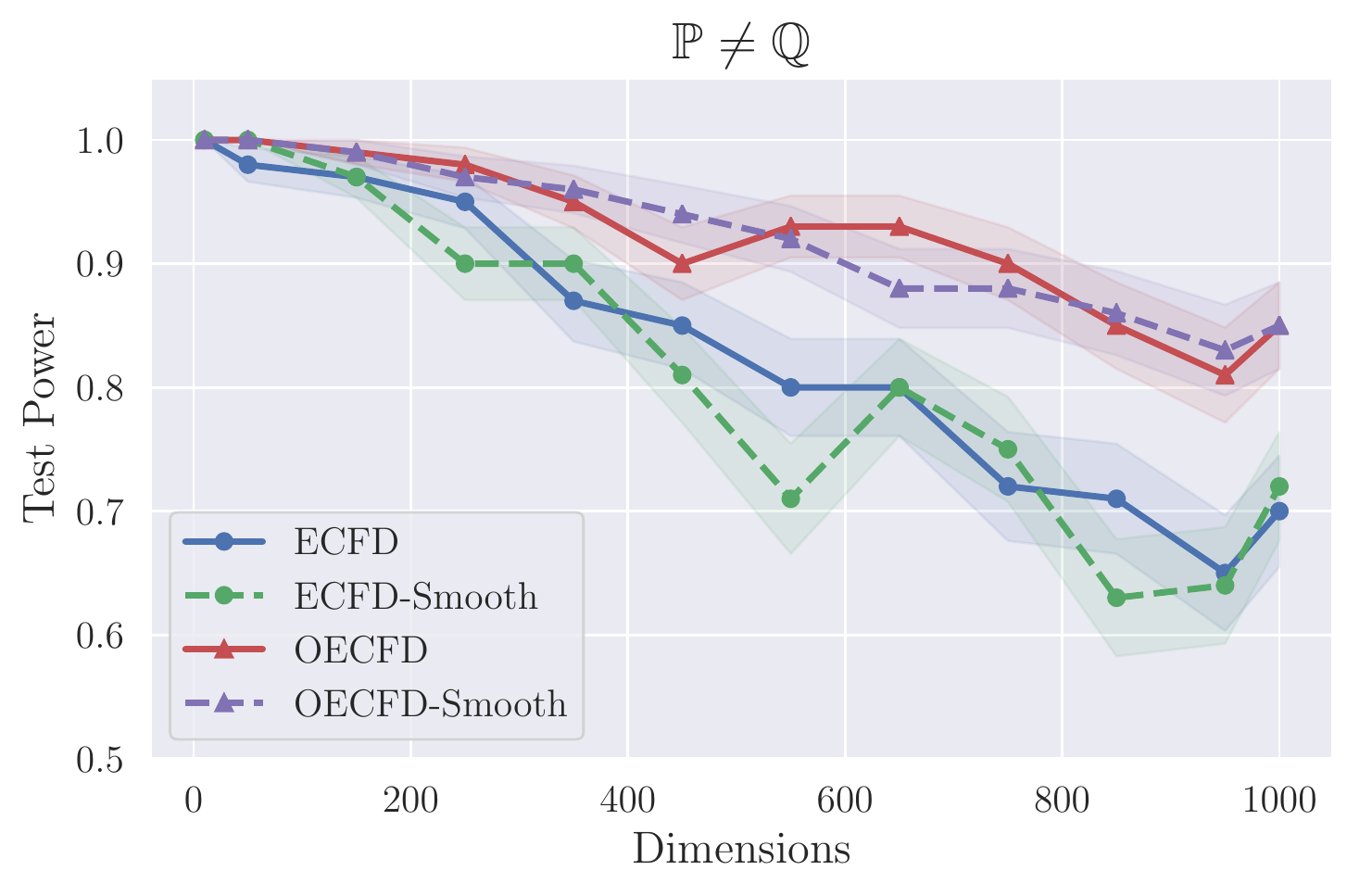}
\label{fig:toyexp1}}\qquad
\subfloat{\includegraphics[width=0.36\linewidth]{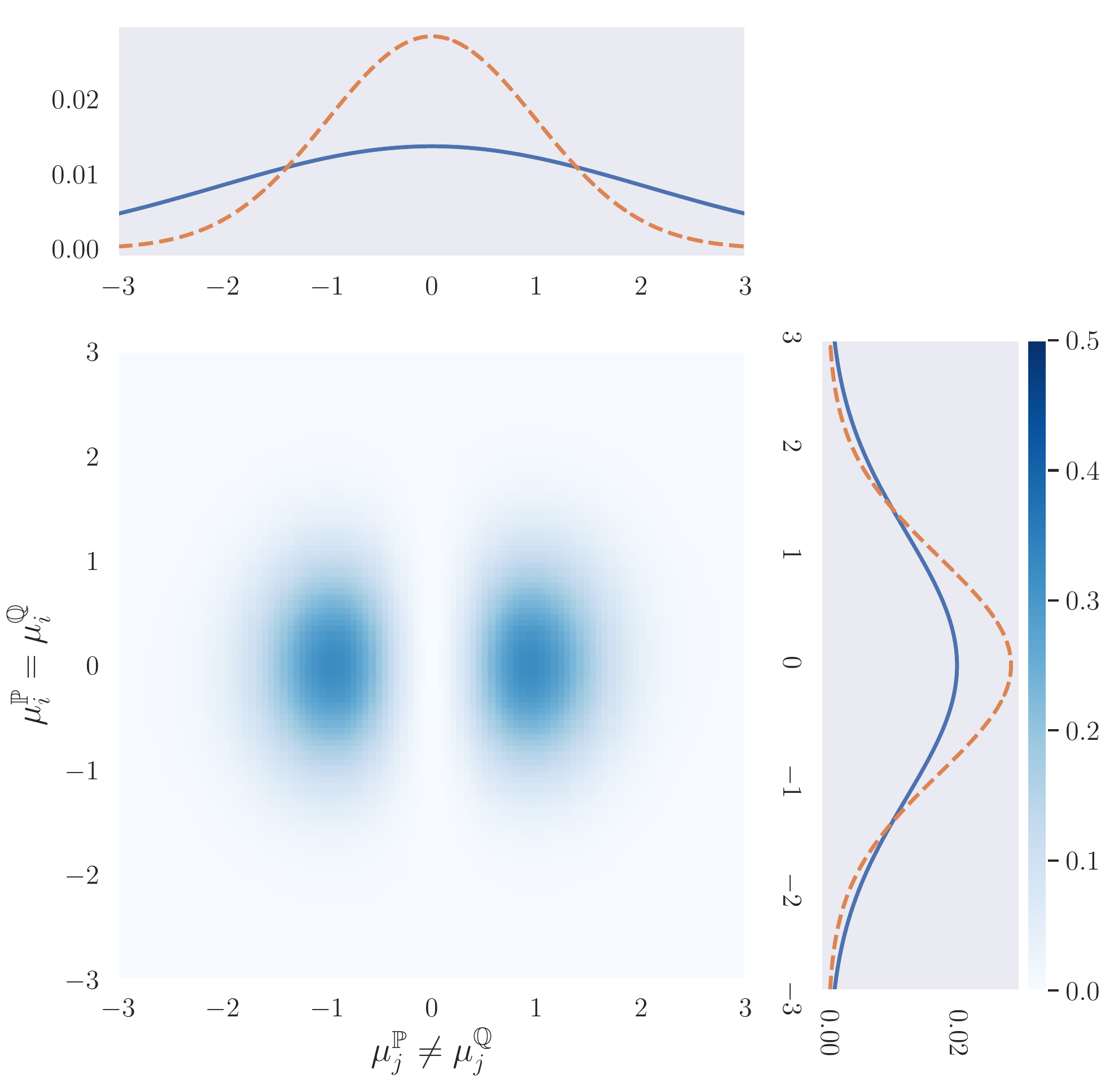}
\label{fig:cfd-visualization}}
\caption{(left) Variation of test power with the number of dimensions for ECFD-based tests; (right) Change in the scale of the weighting distribution upon optimization.}
%\caption{placeholder \hs{Please write up this caption. Can you center (b)? The plot in (a) is rather large and takes up a lot of space. Can we reduce the height and possibly  the range of y. ``test power'' should be ``Test Power'' and ``dimensions'' should be ``Dimensions''. }}
%    \vspace*{-2ex}
\end{figure*}

By analogy to Fourier analysis in signal processing, Eq. (\ref{eq:cfdexpectation}) can be interpreted as the expected discrepancy between the Fourier transforms of two signals at frequencies sampled from $\omega(\mathbf{t};\eta)$. If $\mathrm{supp}(\omega) = \bbR^d$, it can be shown using the uniqueness theorem of characteristic functions that ${\mathrm{CFD}_{\omega}(\mathbb{P},\mathbb{Q})} = 0 \iff \mathbb{P} = \mathbb{Q}$~\cite{sriperumbudur2010hilbert}.

In practice, the CFD can be approximated using empirical characteristic functions and finite samples from the weighting distribution $\omega(\mathbf{t};\eta)$. To elaborate, the characteristic function of a degenerate distribution $\delta_{\mathbf{a}}$ for $\mathbf{a}\in \mathbb{R}^d$ is given by $e^{i\langle \mathbf{t}, \mathbf{a}\rangle}$ where $\mathbf{t}\in \mathbb{R}^d$. Given observations $\mathcal{X} :=\{\mathbf{x}_1,\dots,\mathbf{x}_n\}$ from a probability distribution $\bbP$, the empirical distribution is a mixture of degenerate distributions with equal weights, and the corresponding empirical characteristic function $\hat{\varphi}_{\bbP}$ is a weighted sum of characteristic functions of degenerate distributions:
\begin{align}
\hat{\varphi}_{\bbP}(\mathbf{t}) = \frac{1}{n}\sum_{j=1}^n e^{i\langle \mathbf{t}, \mathbf{x}_j\rangle}.
\end{align}
Let $\mathcal{X} :=\{\mathbf{x}_1,\dots,\mathbf{x}_n\}$ and $\mathcal{Y} :=\{\mathbf{y}_1,\dots,\mathbf{y}_m\}$ with $\mathbf{x}_i,\mathbf{y}_i \in \mathbb{R}^d$ be samples from the distributions $\mathbb{P}$ and $\mathbb{Q}$ respectively, and let $\mathbf{t}_1,\dotsc,\mathbf{t}_k$ be samples from $\omega(\mathbf{t};\eta)$.  
We define the empirical characteristic function distance (ECFD) between $\mathbb{P}$ and $\mathbb{Q}$ as
\begin{align}
{\mathrm{ECFD}_{\omega}^2(\mathbb{P},\mathbb{Q})} &= \frac{1}{k} \sum_{i=1}^k\left|\hat{\varphi}_\mathbb{P}(\mathbf{t}_i) - \hat{\varphi}_\mathbb{Q}(\mathbf{t}_i)\right|^2, \label{eq:ecfdaverage}
\end{align}
where $\hat{\varphi}_\mathbb{P}$ and $\hat{\varphi}_\mathbb{Q}$ are the empirical CFs, computed using $\mathcal{X}$ and $\mathcal{Y}$ respectively. 

A quantity related to CFD (Eq. \ref{eq:cfdint}) has been studied in \cite{paulson1975estimation} and \cite{heathcote1977integrated}, in which the discrepancy between the analytical and empirical characteristic functions of stable distributions is minimized for parameter estimation.  The CFD is well-suited to this application because stable distributions do not admit density functions, making maximum likelihood estimation difficult. Parameter fitting has also been explored for other models such as mixture-of-Gaussians, stable ARMA process, and affine jump diffusion models~\cite{yu2004empirical}.

More recently, \cite{chwialkowski2015fast} proposed fast ($O(n)$ in the number of samples $n$) two-sample tests based on ECFD, as well as a smoothed version of ECFD in which the characteristic function is convolved with an analytic kernel. The authors empirically show that ECFD and its smoothed variant have a better test-power/run-time trade-off compared to quadratic time tests, and better test power than the sub-quadratic time variants of MMD.
 
%\vspace*{-1ex}
\subsection{Optimized ECFD for Two-Sample Testing}
%\vspace*{-1ex}

The choice of $\omega(\mathbf{t};\eta)$ is important for the success of ECFD in distinguishing two different distributions; choosing an appropriate distribution and/or set of parameters $\eta$ allows better coverage of the frequencies at which the differences in $\mathbb{P}$ and $\mathbb{Q}$ lie. For instance, if the differences are concentrated at the frequencies far away from the origin and $\omega(\mathbf{t};\eta)$ is Gaussian, the test power can be improved by suitably enlarging the variance of each coordinate of $\omega(\mathbf{t};\eta)$. 

To increase the power of ECFD, we propose to optimize the parameters $\eta$ (e.g., the variance associated with a normal distribution) of the weighting distribution $\omega(\mathbf{t};\eta)$ to maximize the power of the test. However, care should be taken when specifying how rich the class of functions $\omega(\cdot;\eta)$ is --- the choice of \emph{which} parameters to optimize and the associated constraints is important. Excessive optimization may cause the test to fixate on differences that are merely due to fluctuations in the sampling. As an extreme example, we found that optimizing $\mathbf{t}$'s directly (instead of optimizing the weighting distribution) severely degrades the test's ability to correctly accept the null hypothesis $\mathbb{P} = \mathbb{Q}$.

% care should be taken regarding \emph{which} parameters to optimize and relevant constraints --- allowing unconstrained optimization of all parameters may cause the test to fixate on irrelevant noisy frequencies. Indeed, we found that optimizing $\mathbf{t}$'s directly (instead of optimizing the weighting distribution) severely degrades the test's ability to correctly accept the null hypothesis.  

% this led to an improvement in the ability of the test to distinguish two different distributions but it severely degraded the performance of the test when the two distributions were the same.  
%
%
%It is important to note that while this technique does improve the ability of the test to correctly distinguish two different distributions, it does not hamper its ability to correctly accept the null hypothesis when the distributions are the same (see appendix). Our initial approach was to directly optimize the frequencies at which the distance is measured;
% 

%\hs{I may describe this more generally at first, i.e., state that we can optimize the parameters of the weighting distribution. Again, give the intuition why this would work, e.g., we want to include the frequencies where the differences lie.} 
%\af{If $\omega$ is not normalized, optimizing parameters may be a bad idea. Should not optimize shift parameters because the abs distance between the two CFs is symmetric.}

To validate our approach, we conducted a basic experiment using high-dimensional Gaussians,  similar to~\cite{chwialkowski2015fast}. Specifically, we used two multivariate Gaussians $\mathbb{P}$ and $\mathbb{Q}$ that have the same mean in all dimensions except one. As the dimensionality increases, it becomes increasingly difficult to distinguish between samples from the two distributions. In our tests, the weighting distribution $\omega(\mathbf{t};\eta)$ was chosen to be a Gaussian distribution $\mathcal{N}(\bm{0}, \mathrm{diag}(\bm{\sigma}^2))$, 10000 samples each were taken from $\bbP$ and $\bbQ$, and the number of frequencies ($k$) was set to 3. We optimized the parameter vector $\eta = \{\bm{\sigma}\}$ to maximize the ECFD using the Adam optimizer for 100 iterations with a batch-size of 1000. 

Fig. \ref{fig:toyexp1} shows the variation of the test power (i.e., the fraction of times the null hypothesis $\mathbb{P} = \mathbb{Q}$ is rejected) with the number of dimensions. OEFCD refers to the optimized ECFD, and the ``Smooth'' suffix indicates the smoothed ECFD variant proposed by~\cite{chwialkowski2015fast}. We see that optimization of $\eta$ increases the power of ECFD and ECFD-Smooth, particularly at the higher dimensionalities. There do not appear to be significant differences between the optimized smoothed and non-smoothed ECFD variants. Moreover, the optimization improved the ability of the test to correctly distinguish the two different distributions, but did not hamper its ability to correctly accept the null hypothesis when the distributions are the same (see Appendix \ref{sec:addres}).  

To investigate how $\bm{\sigma}$ is adapted, we visualize two dimensions $\{i,j\}$ from the dataset where $\mu^{\mathbb{P}}_i = \mu^{\mathbb{Q}}_i$ and $\mu^{\mathbb{P}}_j \neq \mu^{\mathbb{Q}}_j$. Fig. \ref{fig:cfd-visualization} shows the absolute difference between the ECFs of $\mathbb{P}$ and $\mathbb{Q}$, with the corresponding dimensions of the weighting distribution plotted in both dimensions. The solid blue line shows the optimized distribution (for OECFD) while the dashed orange line shows the initial distribution (i.e., $\bm{\sigma}=1$  for ECFD and ECFD-Smooth). In the dimension where the distributions are the same, $\sigma$ has small deviation from the initial value. However, in the dimension where the distributions are different, the increase in variance is more pronounced to compensate for the spread of difference between the ECFs away from the origin. 

%\vspace*{-1ex}
\section{Implicit Generative Modeling using CFD}
%\vspace*{-1ex}

In this section, we turn our attention to applying the (optimized) CFD for learning IGMs, specifically GANs. %we view the problem of learning an Implicit Generative Model (IGM) using samples from the data distribution as the problem of matching the characteristic functions of the real and generated data distributions. 
As in the standard GAN, our model is comprised of a generator $g_{\theta}: \calZ \rightarrow \calX$ and a critic $f_{\phi}: \calX \rightarrow \mathbb{R}^m$, with parameter vectors $\theta$ and $\phi$, and data/latent spaces $\calX \subseteq \bbR^d$ and $\calZ \subseteq \bbR^p$.  Below, we write $\Theta,\Phi,\Pi$ for the spaces in which the parameters $\theta,\phi,\eta$ lie.
%The generator takes a latent vector $\mathbf{z} \in \mathcal{Z}$ as input and generates an output $\hat{\mathbf{x}} \in \mathbb{R}^d$ and 

The generator minimizes the empirical CFD between the real and generated data. Instead of minimizing the distance between characteristic functions of raw high-dimensional data, we use a critic neural network $f_\phi$ that is trained to maximize the CFD between real and generated data distributions in a learned lower-dimensional space. This results in the following minimax objective for the IGM:
\begin{align}
	\underset{\theta\in\Theta}{\mathrm{inf}}\:\underset{\psi\in\Psi}{\mathrm{sup}}\:\mathrm{CFD}_{\omega}^2(\mathbb{P}_{f_\phi({\mathcal{X}})}, \mathbb{P}_{f_\phi(g_\theta(\mathcal{Z}))}),
\end{align}
where $\psi = \{\phi, \eta\}$ (with corresponding parameter space $\Psi$), and $\eta$ is the parameter vector of the weighting distribution $\omega$.  The optimization over $\eta$ is omitted if we choose to not optimize the weighting distribution. In our experiments, we set $\eta = \{\bm{\sigma}\}$, with $\bm{\sigma}$ indicating the scale of each dimension of $\omega$. Since evaluating the CFD requires knowledge of the data distribution, in practice, we optimize the empirical estimate $\mathrm{ECFD}_{\omega}^2$ instead of $\mathrm{CFD}_{\omega}^2$. We henceforth refer to this model as the Characteristic Function Generative Adversarial Network (CF-GAN).

%\vspace*{-1ex}
\subsection{CFD Properties: Continuity, Differentiability, and Weak Topology}
%\vspace*{-1ex}

Similar to recently proposed Wasserstein~\cite{Arjovsky2017WassersteinGA} and MMD~\cite{Li2017MMDGT} GANs, the CFD exhibits desirable mathematical properties. Specifically, CFD is continuous and differentiable almost everywhere in the parameters of the generator (Thm. \ref{thm:cont_diff}). Moreover, as it is continuous in the weak topology (Thm. \ref{thm:weak}), it can provide a signal to the generator $g_\theta$ that is more informative for training than other ``distances" that lack this property (e.g., Jensen-Shannon divergence). In the following, we provide proofs for the above claims under assumptions similar to \cite{Arjovsky2017WassersteinGA}.
%Continuity, differentiability, and weak topology are desirable properties for distances/divergences used in GAN critics. When the loss function is discontinuous, it makes optimization using gradient based methods difficult \cite{Arjovsky2017WassersteinGA}. A weak distance can provide a better signal to the generator $g_\theta$ than the one that is not weak. Divergences such as JSD \cite{Goodfellow2014GenerativeAN} are not weak as shown by \cite{Arjovsky2017TowardsPM}. More recently used distances for training GANs such as Wasserstein \cite{Arjovsky2017WassersteinGA}, MMD \cite{Li2017MMDGT}, etc., follow the aforementioned properties under mild assumptions. In the following, we show that CFD exhibits these properties under similar assumptions.
%\begin{assumption}
%	$g_\theta: \mathcal{Z} \rightarrow \mathcal{X}$ is locally Lipschitz, $f_\phi$ is Lipschitz, and $f_\phi \circ g_\theta$ is differentiable in $\theta$. There are local Lipschitz constants $L(\theta, \mathbf{z})$ for $f_\phi \circ g_\theta$, independent of $\phi$, such that $\mathbb{E}_{\mathbf{z}}\left[L(\theta, \mathbf{z})\right] < \infty$.
%	\label{assump:1}
%\end{assumption}
%\begin{assumption}
%	For all $\eta \in \Pi$, the weighting distribution $\omega(\mathbf{t})$ has support $\bbR^m$, where $m$ is the output dimensionality of $f_\phi$. In addition, $\sup_{\eta \in \Pi} \mathbb{E}_{\omega(\mathbf{t})}\left[\|\mathbf{t}\|\right] < \infty$, where $\omega(\cdot)$ implicitly depends on $\eta$.
%	\label{assump:2}
%\end{assumption}

The following theorem formally states the result of continuity and differentiability in $\theta$ almost everywhere, which is desirable for permitting training via gradient descent. 

\begin{theorem} \label{thm:cont_diff}
    Assume that (i) $f_\phi \circ g_\theta$ is locally Lipschitz with respect to $(\theta, \mathbf{z})$ with constants $L(\theta, \mathbf{z})$ not depending on $\phi$ and satisfying $\mathbb{E}_{\mathbf{z}}\left[L(\theta, \mathbf{z})\right] < \infty$; (ii) $\sup_{\eta \in \Pi} \mathbb{E}_{\omega(\mathbf{t};\eta)}\left[\|\mathbf{t}\|\right] < \infty$. Then, the function~$\mathrm{sup}_{\psi \in \Psi}\mathrm{CFD}_{\omega}^2(\mathbb{P}_{f_\phi({\mathcal{X}})}, \mathbb{P}_{f_\phi(g_\theta(\mathcal{Z}))})$ is continuous in $\theta \in \Theta$ everywhere, and differentiable in $\theta \in \Theta$ almost everywhere.
    % Under Assumptions \ref{assump:1} and \ref{assump:2}, 
%    Then, the function~$\underset{\psi \in \Psi}{\mathrm{sup}}\:\mathrm{CFD}_{\omega}^2(\mathbb{P}_{f_\phi({\mathcal{X}})}, \mathbb{P}_{f_\phi(g_\theta(\mathcal{Z}))})$ is continuous in $\theta \in \Theta$ everywhere, and differentiable in $\theta \in \Theta$ almost everywhere.
\end{theorem}
\vspace*{-0.5ex}
The following theorem establishes continuity in the weak topology, and concerns general convergent distributions as opposed to only those corresponding to $g_{\theta}(\mathbf{z})$.  In this result, we let $\mathbb{P}^{(\phi)}$ be the distribution of $f_{\phi}(\mathbf{x})$ when $\mathbf{x} \sim \mathbb{P}$, and similarly for $\mathbb{P}_{n}^{(\phi)}$. 

\begin{theorem} \label{thm:weak}
    Assume that (i) $f_\phi$ is $L_f$-Lipschitz for some $L_f$ not depending on $\phi$; (ii) $\sup_{\eta \in \Pi} \mathbb{E}_{\omega(\mathbf{t})}\left[\|\mathbf{t}\|\right] < \infty$.
    % Under Assumptions \ref{assump:1} and \ref{assump:2}, 
    Then, the function~$\mathrm{sup}_{\psi \in \Psi}\mathrm{CFD}_{\omega}^2(\mathbb{P}_{n}^{(\phi)},\mathbb{P}^{(\phi)})$ is continuous in the weak topology, i.e., if $\mathbb{P}_n \xrightarrow{D} \mathbb{P}$, then $\mathrm{sup}_{\psi \in \Psi}\mathrm{CFD}_{\omega}^2(\mathbb{P}_{n}^{(\phi)},\mathbb{P}^{(\phi)}) \rightarrow 0$, where $\xrightarrow{D}$ implies convergence in distribution.
\end{theorem}
\vspace*{-0.5ex}
The proofs are given in the appendix. In brief, we bound the difference between characteristic functions using geometric arguments; we interpret $e^{ia}$ as a vector on a circle, and note that $|e^{ia} - e^{ib}| \le |a-b|$.  We then upper-bound the difference of function values in terms of $\mathbb{E}_{ \omega(\mathbf{t}) }[ \|\mathbf{t}\|]$ (assumed to be finite) and averages of Lipschitz functions of $\mathbf{x}, \mathbf{x}'$ under the distributions  considered. The Lipschitz properties ensure that the function difference vanishes when one  distribution converges to the other.

Various generators satisfy the locally Lipschitz assumption, e.g., when  $g_\theta$ is a feed-forward network with ReLU activations. To ensure that $f_\phi$ is Lipschitz, common methods employed in prior work include weight clipping \cite{Arjovsky2017WassersteinGA} and gradient penalty \cite{Gulrajani2017ImprovedTO}.
In addition, many common distributions satisfy $\mathbb{E}_{\omega(\mathbf{t})}\left[\|\mathbf{t}\|\right] < \infty$ , e.g., Gaussian, Student-t, and Laplace with fixed $\bm{\sigma}$. When $\bm{\sigma}$ is unbounded and optimized, we normalize the CFD by $\|\bm{\sigma}\|$, which prevents $\bm{\sigma}$ from going to infinity.

An example demonstrating the necessity of Lipschitz assumptions in continuity results (albeit for a different metric) can be found in Example 1 of \cite{arbel2018gradient}.
In the appendix, we discuss conditions under which Theorem \ref{thm:weak} can be strengthened to an ``if and only if'' statement.

%\hs{theorems come here. Should we put everything into one Theorem above or in two?}

%Our experiments are conducted with these three distributions, but we include the uniform distribution as an example when this assumption is violated. 

%A number of distributions have support $\bbR^m$. We conduct experiments with Gaussian, Student's-t, and Laplace distributions which all satisfy this property. We also conduct experiments with the uniform distribution to see what happens when this condition is broken. For a fixed value of scale parameter of all these distributions, the condition $\mathbb{E}_{\omega(\mathbf{t})}\left[\|\mathbf{t}\|\right] < \infty$ holds. For the case when $\bm{\sigma}$ is optimized to maximize CFD, we normalize the CFD by $\|\bm{\sigma}\|$ which prevents the elements of $\bm{\sigma}$ from exploding. 
%\begin{theorem}
%	$\underset{\phi}{\mathrm{max}}\:\mathrm{CFD}_{\omega}^2(\mathbb{P}_{f_\phi({\mathcal{X}})}, \mathbb{P}_{f_\phi(g_\theta(\mathcal{Z}))})$ is continuous everywhere, differentiable almost everywhere in $\theta$, and is weak, i.e., if $\mathbb{P}_n \xrightarrow{D} \mathbb{P}$, then $\underset{\phi}{\mathrm{max}}\:\mathrm{CFD}_{\omega}^2(\mathbb{P}_{n},\mathbb{P}) \rightarrow 0$, where $\xrightarrow{D}$ implies convergence in distribution.
%\end{theorem}
%\begin{proof}
%	(See appendix).
%\end{proof}
% Feasible set reduction

%\section{Related Work}
%\paragraph{Characteristic Function} 

%\vspace*{-1ex}
\subsection{Relation to MMD and Prior Work}
%\vspace*{-1ex}

The CFD is related to the maximum mean discrepancy (MMD)~\cite{gretton2012kernel}. 
Given samples from two distributions $\mathbb{P}$ and $\mathbb{Q}$, the squared MMD is given by 
\begin{align}
\mathrm{MMD}_{k}^2(\mathbb{P}, \mathbb{Q}) = \mathbb{E}\left[\kappa(x,x')\right] &+ \mathbb{E}\left[\kappa(y,y')\right]-2\mathbb{E}\left[\kappa(x,y)\right]
\end{align}
where $x, x' \sim \bbP$ and $y, y' \sim \bbQ$ are independent samples, and $\kappa$ is kernel. When the weighting distribution of the CFD is equal to the inverse Fourier transform of the kernel in MMD (i.e., $\omega(\mathbf{t}) = \mathcal{F}^{-1}\left\{\kappa\right\}$), the CFD and squared MMD are equivalent: ${\mathrm{CFD}_{\omega}^2(\mathbb{P},\mathbb{Q})} = \mathrm{MMD}^2_{\kappa}(\mathbb{P}, \mathbb{Q})$. 
Indeed, kernels with $\mathrm{supp}(\mathcal{F}^{-1}(\kappa)) = \bbR^d$ are called \textit{characteristic kernels} \cite{sriperumbudur2010hilbert}, and when $\mathrm{supp}(\omega) = \bbR^d$, $\mathrm{MMD}_{\kappa}(\mathbb{P}, \mathbb{Q}) = 0$ if and only if $\mathbb{P} = \mathbb{Q}$. Although formally equivalent under the above conditions, we find experimentally that optimizing empirical estimates of MMD and CFD result in different convergence profiles and model performance across a range of datasets. Also, unlike $\textrm{MMD}$, which takes quadratic time in the number of samples to approximately compute, the CFD takes $O(nk)$ time and is therefore computationally attractive when $k \ll n$. 

Learning a generative model by minimizing the MMD between real and generated samples was proposed independently by \cite{Li2015GenerativeMM} and \cite{Dziugaite2015TrainingGN}. The Generative Moment Matching Network (GMMN) \cite{Li2015GenerativeMM} uses an autoencoder to first transform the data into a latent space, and then trains a generative network to produce latent vectors that match the true latent distribution. The MMD-GAN \cite{Li2017MMDGT} performs a similar input transformation using a network $f_\phi$ that is adversarially trained to maximize the MMD between the true distribution $\mathbb{P}_{\mathcal{X}}$ and the generator distribution $\mathbb{Q}_{\theta}$; this results in a GAN-like min-max criterion. More recently, \cite{Binkowski2018DemystifyingMG} and \cite{arbel2018gradient} have proposed different theoretically-motivated regularizers on the gradient of MMD-GAN critic that improve training. In our experiments, we compare against the MMD-GAN both with and without gradient regularization. 

Very recent work~\cite{Li2019ImplicitKL} (IKL-GAN) has evaluated kernels parameterized in Fourier space, which are then used to compute MMD in MMD-GAN. In contrast to IKL-GAN, we derive the CF-GAN via characteristic functions rather than via MMD, and our method obviates the need for kernel evaluation. We also provide novel direct proofs for the theoretical properties of the optimized CFD that are \emph{not} based on its equivalence to MMD. The IKL-GAN utilizes a neural network to sample random frequencies, whereas we use a simpler fixed distribution with a learned scale, reducing the number of hyperparameters to tune. Our method yields state-of-the-art performance, which suggests that the more complex setup in IKL-GAN may not be required for effective GAN training. 

In parallel, significant work has gone into improving GAN training via architectural and optimization enhancements \cite{miyato2018spectral,brock2018large,karras2018style}; these research directions are orthogonal to our work and can be incorporated in our proposed model.

\begin{figure*}
	\centering
	\subfloat{\includegraphics[width=0.40\linewidth]{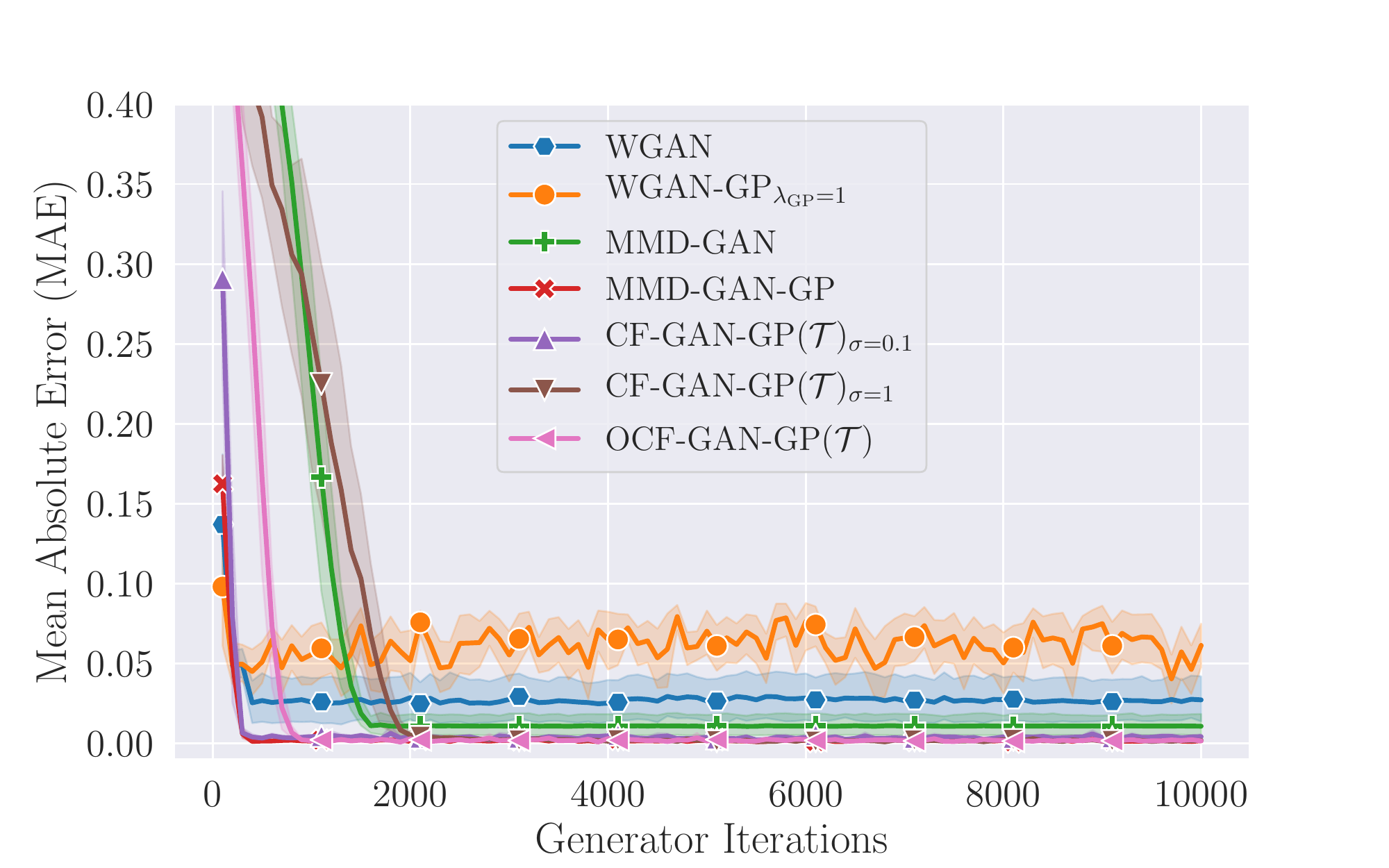}
		\label{fig:synth-uni}}
	\subfloat{\includegraphics[width=0.40\linewidth]{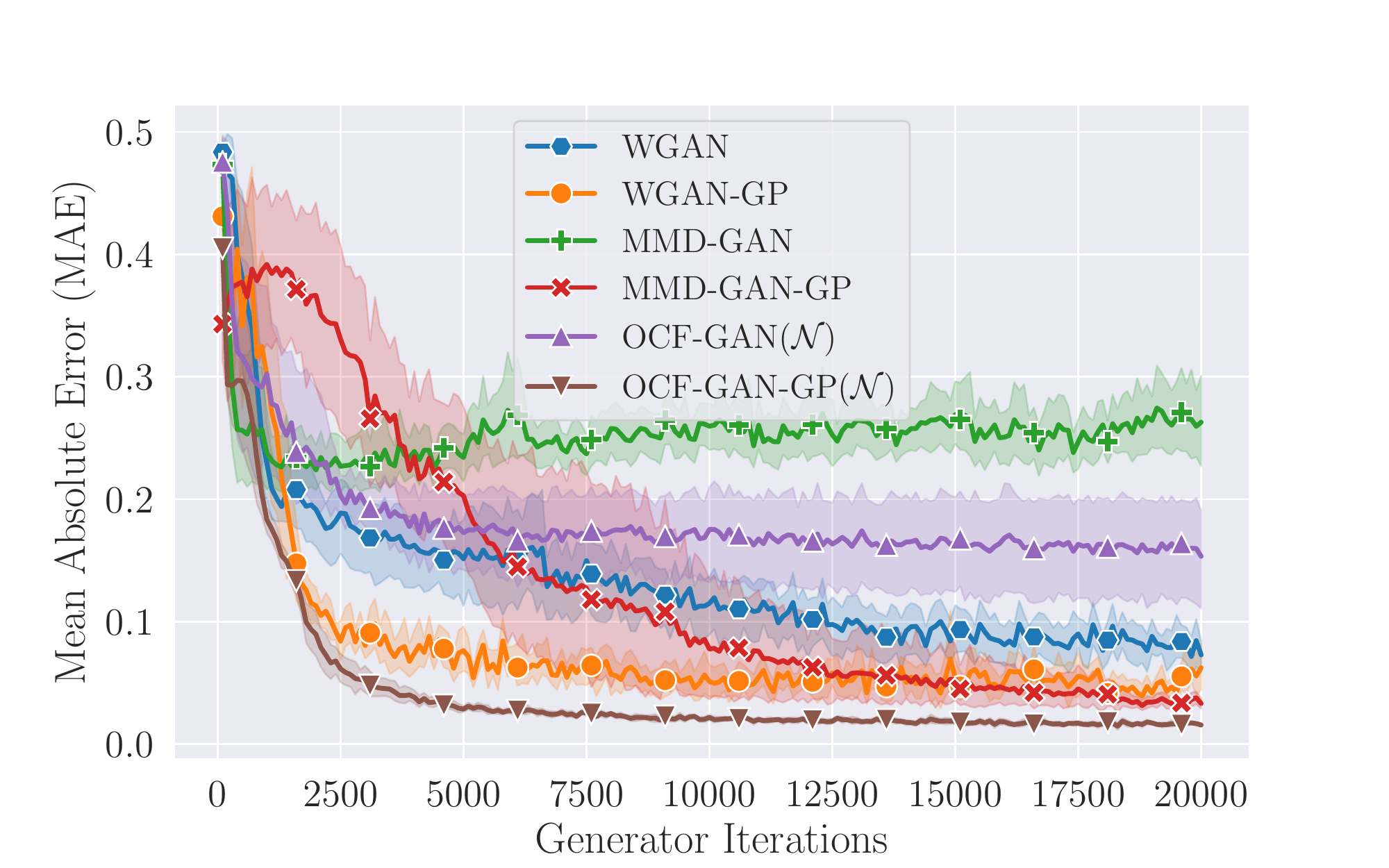}
		\label{fig:synth-bi}}
	\caption{Variation of MAE for synthetic datasets $\mathcal{D}_1$ (left) and $\mathcal{D}_2$ (right) with generator iterations. The plots are averaged over 10 random runs.}
%    \vspace*{-2ex}
\end{figure*}

%\vspace*{-1ex}
\section{Experiments}
%\vspace*{-1ex}

In this section, we present empirical results comparing different variants of our proposed model: CF-GAN. We prefix O to the model name when the $\bm{\sigma}$ parameters were optimized along with the critic and omit it when $\bm{\sigma}$ was kept fixed. Similarly, we suffix GP to the model name when gradient penalty~\cite{Gulrajani2017ImprovedTO} was used to enforce Lipschitzness of $f_\phi$. In the absence of gradient penalty, we clipped the weights of $f_\phi$ in $[-0.01, 0.01]$. When the parameters $\bm{\sigma}$ were optimized, we scaled the ECFD by $\|\bm{\sigma}\|$ to prevent $\bm{\sigma}$ from going to infinity, thereby ensuring $\mathbb{E}_{\omega(\mathbf{t})}\left[\|\mathbf{t}\|\right] < \infty$.
%In this section, we present empirical results comparing three variants of our proposed model: 1) CF-GAN -- ECFD with fixed scale $\bm{\sigma}$; 2) OCF-GAN -- ECFD with optimized $\bm{\sigma}$; and, 3) OCF-GAN-GP -- ECFD with optimized $\bm{\sigma}$ and gradient penalty. For the first two versions, we use weight clipping in $[-0.01, 0.01]$ to enforce Lipschitzness of $f_\phi$, while for OCF-GAN-GP we constrain the gradients between real and generated samples to be 1 using an additive penalty~\cite{Gulrajani2017ImprovedTO}. 

We compare our proposed model against two variants of MMD-GAN: (i) MMD-GAN \cite{Li2017MMDGT}, which uses MMD with a mixture of RBF kernels as the distance metric; (ii) MMD-GAN-GP$_{L2}$ \cite{Binkowski2018DemystifyingMG}, which introduces an additive gradient penalty based on MMD's IPM witness function, an L2 penalty on discriminator activations, and uses a mixture of RQ kernels. We also compare against WGAN \cite{Arjovsky2017WassersteinGA} and WGAN-GP \cite{Gulrajani2017ImprovedTO} due to their close relation to MMD-GAN~\cite{Li2017MMDGT,Binkowski2018DemystifyingMG}. Our code is available online at \href{https://github.com/crslab/OCFGAN}{https://github.com/crslab/OCFGAN}.

%\vspace*{-1ex}
\subsection{Synthetic Data} 
%\vspace*{-1ex}

We first tested the methods on two synthetic 1D distributions: a simple unimodal distribution ($\mathcal{D}_1$) and a more complex bimodal distribution ($\mathcal{D}_2$). The distributions were constructed by transforming $z \sim \calN(0,1)$ using a function $h: \bbR \to \bbR$. For the unimodal dataset, we used the scale-shift function form used by \cite{Zaheer2018ConnoisseurC}, where $h(z) = \mu + \sigma z$. For the bimodal dataset, we used the function form used by planar flow \cite{rezende2015variational}, where $h(z) = \alpha z + \beta\tanh(\gamma\alpha z)$. We trained the various GAN models to approximate the distribution of the transformed samples. Once trained, we compared the transformation function $\hat{h}$ learned by the GAN against the true function $h$. We computed the mean absolute error (MAE) ($\mathbb{E}_{z}[|h(z) - \hat{h}(z)|]$) to evaluate the models. Further details on the experimental setup are in Appendix \ref{sec:synthetic}.

Figs.~\ref{fig:synth-uni} and \ref{fig:synth-bi} show the variation of the MAE with training iterations. For both datasets, the models with gradient penalty converge to better minima. In $\mathcal{D}_1$, MMD-GAN-GP and OCF-GAN-GP converge to the same value of MAE, but MMD-GAN-GP converges faster. During our experiments, we observed that the scale of the weighting distribution (which is intialized to 1) falls rapidly before the MAE begins to decrease. For the experiments with the scale fixed at 0.1 (CF-GAN-GP$_{\sigma=0.1}$) and 1 (CF-GAN-GP$_{\sigma=1}$), both models converge to the same MAE, but CF-GAN-GP$_{\sigma=1}$ takes much longer to converge than CF-GAN-GP$_{\sigma=0.1}$. This indicates that the optimization of the scale parameter leads to faster convergence. For the more complex dataset $\mathcal{D}_2$, MMD-GAN-GP takes significantly longer to converge compared to WGAN-GP and OCF-GAN-GP. OCF-GAN-GP converges fastest and to a better minimum, followed by WGAN-GP. 

\begin{figure*}[htb]
	\centering
	\subfloat{\includegraphics[width=0.28\textwidth]{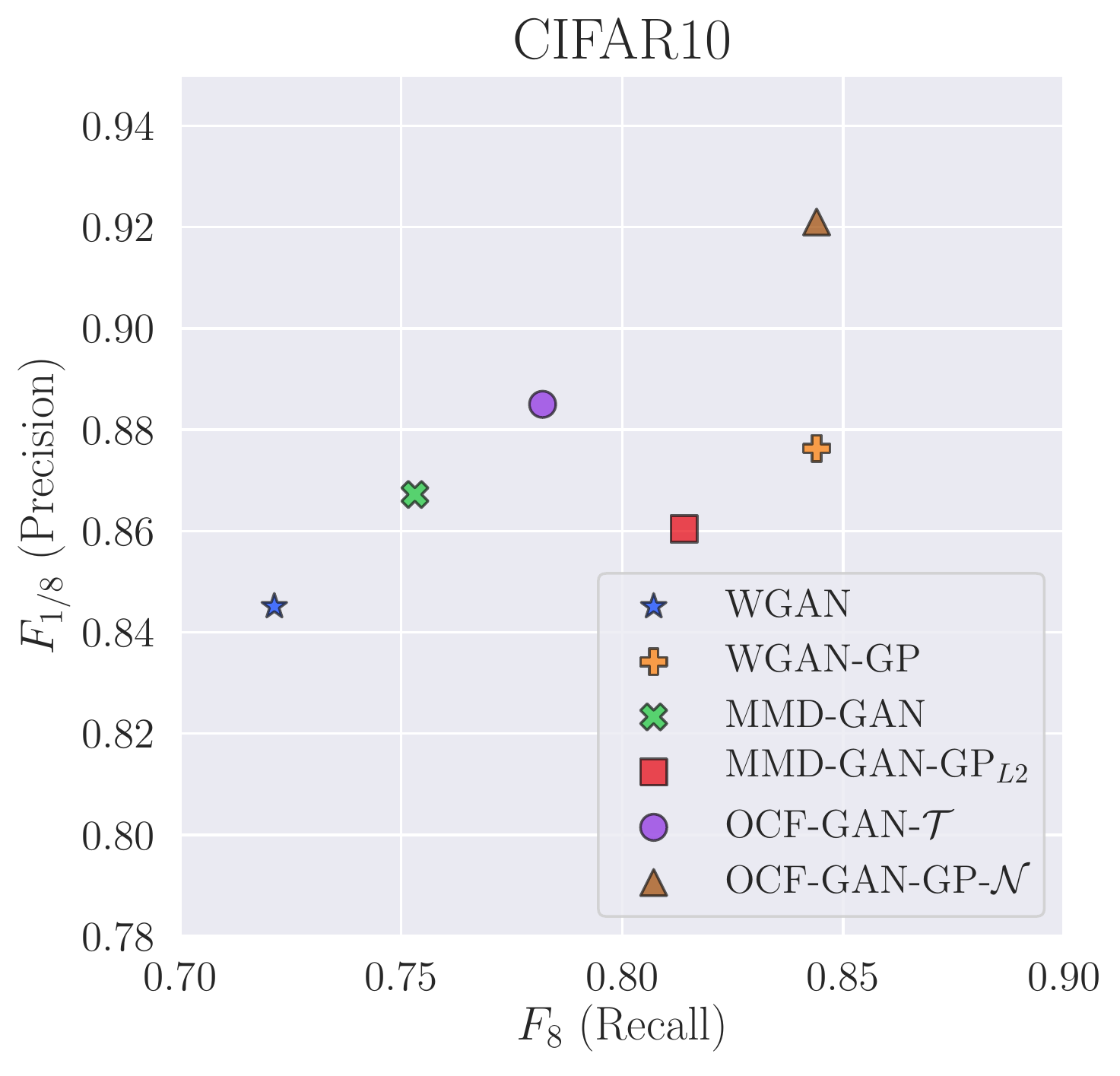}}\quad
	\subfloat{\includegraphics[width=0.28\textwidth]{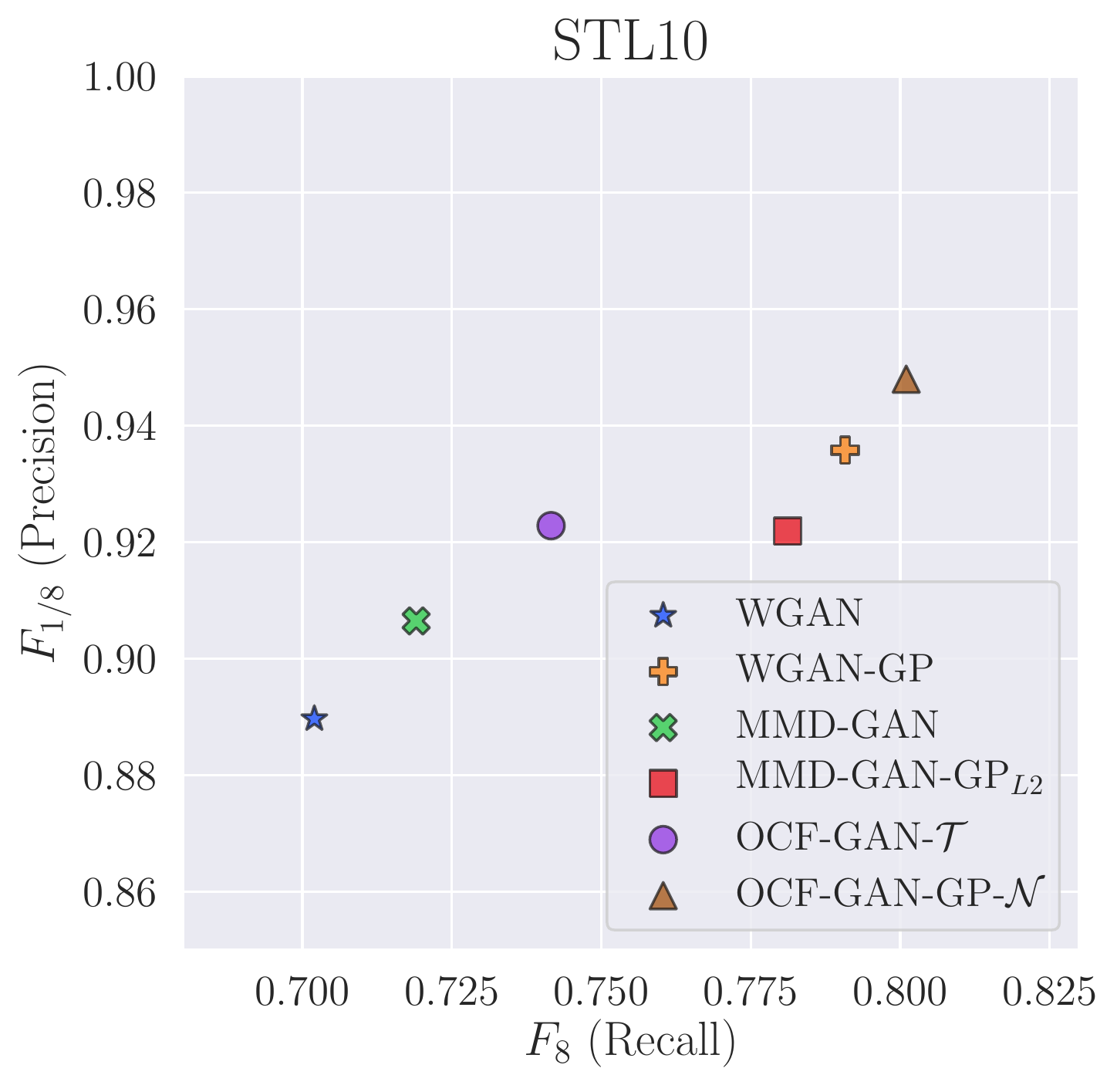}}
	\quad
	\subfloat{\includegraphics[width=0.28\textwidth]{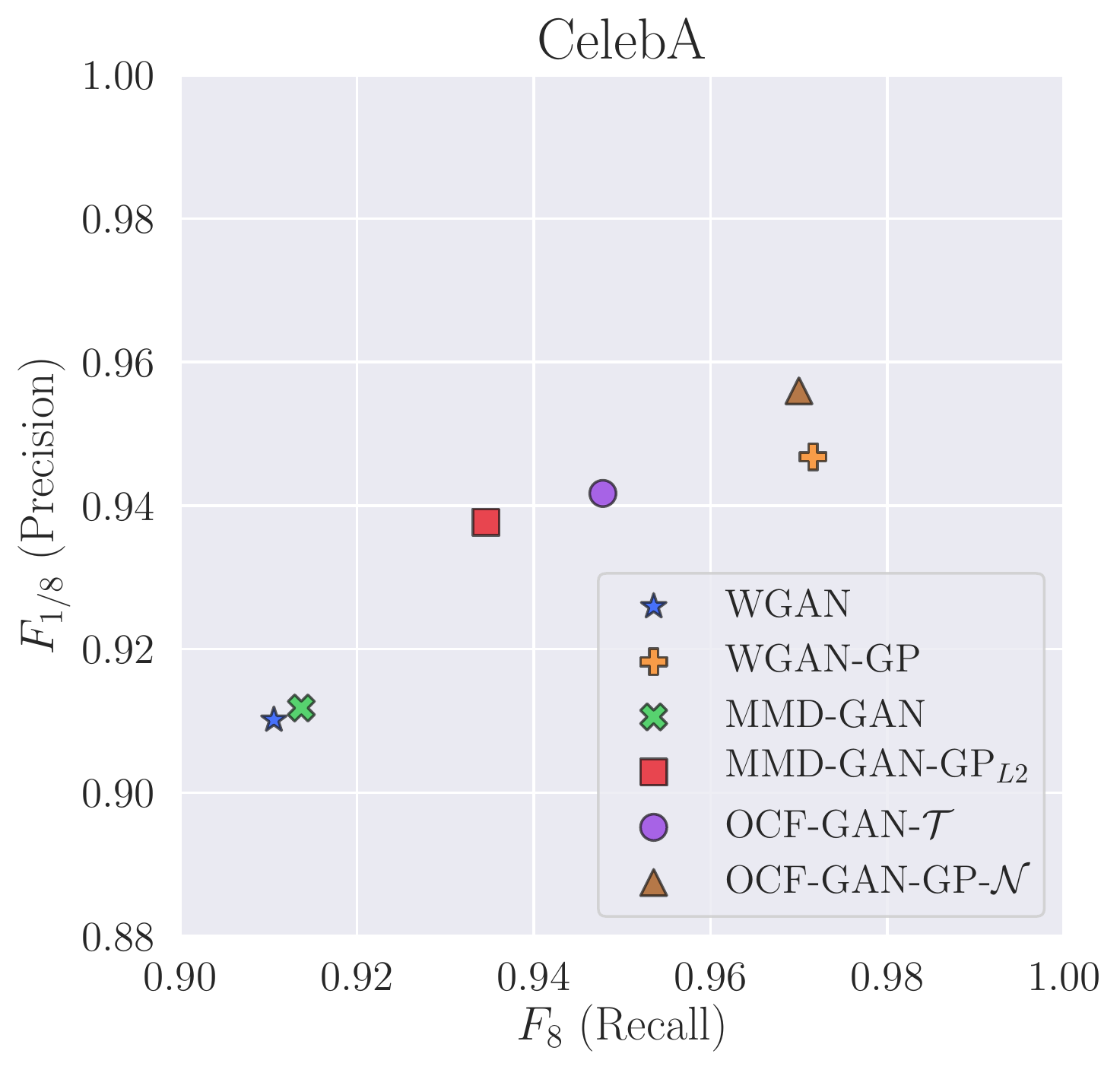}}
	\caption{Precision-Recall scores (higher is better) for CIFAR10 (left), STL10 (center), and CelebA (right) datasets.}
	\label{fig:prplot}
%    \vspace*{-3ex}
\end{figure*}

%\vspace*{-1ex}
\subsection{Image Generation}
%\vspace*{-1ex}

A recent large-scale analysis of GANs~\cite{lucic2018gans} showed that different models achieve similar \emph{best} performance when given \emph{ample} computational budget, and advocates comparisons between distributions under practical settings. As such, we compare scores attained by the models from different initializations under fixed computational budgets. We used four datasets: 1) MNIST \cite{LeCun2001GradientbasedLA}: 60K grayscale images of handwritten digits; 2) CIFAR10 \cite{Krizhevsky2009LearningML}: 50K RGB images; 3) CelebA \cite{liu2015faceattributes}: $\approx$200K RGB images of celebrity faces; and 4) STL10 \cite{coates2011analysis}: 100K RGB images. For all datasets, we center-cropped and scaled the images to $32 \times 32$. 

%\vspace*{-2ex}
\paragraph{Network and Hyperparameter Details} Given our computational budget and experiment setup, we used a DCGAN-like generator $g_\theta$ and critic $f_\phi$ architecture for all models (similar to \cite{Li2017MMDGT}). For MMD-GAN, we used a mixture of five RBF kernels (5-RBF) with different scales~\cite{Li2017MMDGT}. MMD-GAN-GP$_{L2}$ used a mixture of rational quadratic kernels (5-RQ). The kernel parameters and the trade-off parameters for gradient and L2 penalties were set according to \cite{Binkowski2018DemystifyingMG}. We tested CF-GAN variants with two weighting distributions: Gaussian ($\mathcal{N}$) and Student's-t ($\mathcal{T}$) (with 2 degrees of freedom). For CF-GAN, we tested 3 scale parameters in the set $\{0.2, 0.5, 1\}$, and we report the best results. The number of frequencies ($k$) for computing ECFD was set to 8. Please see Appendix \ref{sec:imagexp} for implementation details.
\paragraph{Evaluation Metrics} We compare the different models using three evaluation metrics: Fr\'echet Inception Distance (FID)~\cite{salimans2016improved}, Kernel Inception Distance (KID)~\cite{Binkowski2018DemystifyingMG}, and Precision-Recall (PR) for generative models~\cite{sajjadi2018assessing}. Details on these metrics and the evaluation procedure can be found in Appendix \ref{sec:imagexp}. In brief, the FID computes the Fr\'echet distance between two multivariate Gaussians and the KID computes the MMD (with a polynomial kernel of degree 3) between the real and generated data distributions. Both FID and KID give single value scores, and PR gives a two dimensional score which disentangles the quality of generated samples from the coverage of the data distribution. PR is defined by a pair $F_8$ (recall) and $F_{1/8}$ (precision) which represent the coverage and sample quality, respectively~\cite{sajjadi2018assessing}. 
\paragraph{Results} In the following, we summarize our main findings, and relegate the details to the Appendix. 
Table \ref{tab:cifar} shows the FID and KID values achieved by different models for CIFAR10, STL10, and CelebA datasets. In short, our model outperforms both variants of WGAN and MMD-GAN by a significant margin. OCF-GAN, using just one weighting function, outperforms both MMD-GANs that use a mixture of 5 different kernels. 

We observe that the optimization of the scale parameter improves the performance of the models for both weighting distributions, and the introduction of gradient penalty as a means to ensure Lipschitzness of $f_\phi$ results in a significant improvement in the score values for all models. This is in line with the results of \cite{Gulrajani2017ImprovedTO} and \cite{Binkowski2018DemystifyingMG}. Overall, amongst the CF-GAN variants, OCF-GAN-GP with Gaussian weighting performs the best for all datasets. 

%\vspace*{-1ex}
The two-dimensional precision-recall scores in Fig.~\ref{fig:prplot} provide further insight into the performance of different models. Across all the datasets, the addition of gradient penalty (OCF-GAN-GP) rather than weight clipping (OCF-GAN) leads to a higher improvement in recall compared to precision. %Recall measures how well the generative model covers the real data distribution. 
This result supports recent arguments that weight clipping forces the generator to learn simpler functions, while gradient penalty is more flexible~\cite{Gulrajani2017ImprovedTO}. The improvement in recall with the introduction of gradient penalty is more noticeable for CIFAR10 and STL10 datasets compared to CelebA. This result is intuitive; CelebA is a more uniform and simpler dataset compared to CIFAR10/STL10, which contain more diverse classes of images, and thus likely have modes that are more complex and far apart. Results on the MNIST dataset, where all models achieve good score values, are available in Appendix \ref{sec:addres}, which also includes further experiments using the smoothed version of ECFD and the optimized smoothed version (no improvement over the unsmoothed versions on the image datasets).

\begin{figure}
	\centering
	\subfloat{\includegraphics[width=0.75\columnwidth]{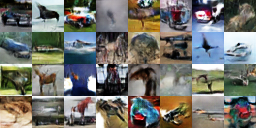}}\\
	\subfloat{\includegraphics[width=0.75\columnwidth]{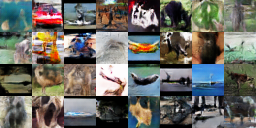}}
	\\
%	\subfloat{\includegraphics[width=0.75\columnwidth]{celeba-ocfgangp-main}}\\
	\subfloat{\includegraphics[width=0.75\columnwidth]{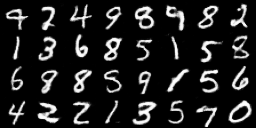}}
	\caption{Image samples for the different datasets (top to bottom: CIFAR10, STL10, and MNIST) generated by OCF-GAN-GP (random samples without selection).}
	\label{fig:imagesamples}
%    \vspace*{-2ex}
\end{figure}

\begin{figure}
	\centering
	\subfloat{\includegraphics[width=0.9\columnwidth]{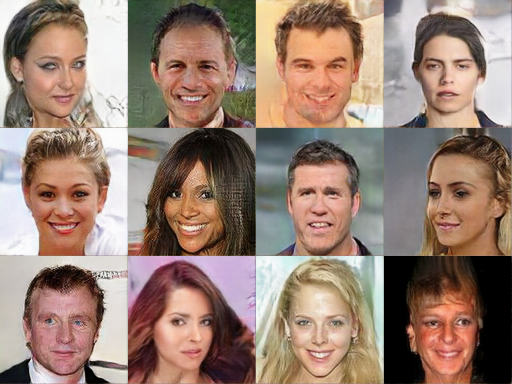}}
	\caption{Image samples for the $128 \times 128$ CelebA dataset generated by OCF-GAN-GP with a ResNet generator (random samples without selection).}
	\label{fig:resnetsamples}
%    \vspace*{-2ex}
\end{figure}

\paragraph{Qualitative Results} In addition to the quantitative metrics presented above, we also performed a qualitative analysis of the generated samples. Fig. \ref{fig:imagesamples} shows image samples generated by OCF-GAN-GP for different datasets. We also tested our method with a deep ResNet model on a $128 \times 128$ scaled version of CelebA dataset. Samples generated by this model (Fig. \ref{fig:resnetsamples}) show that OCF-GAN-GP scales to larger images and networks, and is able to generate visually appealing images  comparable to state-of-the-art methods using similar sized networks. Additional qualitative comparisons can be found in Appendix \ref{sec:addres}.

\begin{table*}
	\caption{FID and KID ($\times 10^3$) scores (lower is better) for CIFAR10, STL10, and CelebA datasets averaged over 5 random runs (standard deviation in parentheses).}
	\label{tab:cifar}
	\centering
	\begin{tabular}{llrrrrrr} 
		\toprule
		\multirow{2}{*}{Model}                        & \multirow{2}{*}{Kernel/}  & \multicolumn{2}{c}{CIFAR10} & \multicolumn{2}{c}{STL10} & \multicolumn{2}{c}{CelebA} \\ 
		\cmidrule{3-8}
		&Weight                       & \multicolumn{1}{c}{FID}          & \multicolumn{1}{c}{KID}          & \multicolumn{1}{c}{FID}          & \multicolumn{1}{c}{KID}   &  \multicolumn{1}{c}{FID}          & \multicolumn{1}{c}{KID}         \\ 
		\midrule
		WGAN               & –                           & 44.11 (1.16) & 25 (1)       & 38.61 (0.43) & 23 (1)  & 17.85 (0.69) & 12 (1)      \\
		WGAN-GP            & –                           & 35.91 (0.30) & 19 (1)       & 27.85 (0.81) & 15 (1)  & 10.03 (0.37) & 6 (1)    \\
		MMD-GAN            & 5-RBF                       & 41.28 (0.54) & 23 (1)       & 35.76 (0.54) & 21 (1)  & 18.48 (1.60) & 12 (1)    \\
		MMD-GAN-GP$_{L2}$      & 5-RQ                        & 38.88 (1.35) & 21 (1)       & 31.67 (0.94) & 17 (1)  & 13.22 (1.30) & 8 (1) \\ 
		\midrule
		\multirow{2}{*}{CF-GAN}
		& $\mathcal{N}_{(\sigma=0.5)}$    & 39.81 (0.93) & 23 (1)       & 33.54 (1.11) & 19 (1)  & 13.71 (0.50) & 9 (1)    \\
		& $\mathcal{T}_{(\sigma=1)}$      & 41.41 (0.64) & 22 (1)       & 35.64 (0.44) & 20 (1)  & 16.92 (1.29) & 11 (1)    \\ 
		\cmidrule{2-8}
		\multirow{2}{*}{OCF-GAN}
		& $\mathcal{N}$  & 38.47 (1.00) & 20 (1)       & 32.51 (0.87) & 19 (1)  & 14.91 (0.83) & 9 (1)    \\
		& $\mathcal{T}$  & 37.96 (0.74) & 20 (1)       & 31.03 (0.82) & 17 (1)  & 13.73 (0.56) & 8 (1)    \\
		\cmidrule{2-8}
		\multirow{2}{*}{OCF-GAN-GP}
		& $\mathcal{N}$  & \textbf{33.08 (0.26)}  &  \textbf{17 (1)}  & \textbf{26.16 (0.64)} &   \textbf{14 (1)}  & \textbf{9.39 (0.25)}  & \textbf{5 (1)}  \\
		& $\mathcal{T}$  & 34.33 (0.77)  &  18 (1)  & 26.86 (0.38) &   15 (1)  & 9.61 (0.39)  &  6 (1)   \\
		\bottomrule
	\end{tabular}
%	   \vspace*{-3ex}
\end{table*}
\begin{table}
	\caption{FID and KID scores for on the MNIST dataset with varying numbers of frequencies used in OCF-GAN-GP.}
	\label{tab:freqexp}
	\centering
	\begin{tabular}{crr}
		\toprule
		\# of freqs ($k$) & \multicolumn{1}{c}{FID} & \multicolumn{1}{c}{KID $\times 10^3$}\\
		\midrule
		1 & 0.44 (0.03) & 5 (1)\\
		4 & 0.39 (0.05) & 4 (1)\\
		8 & 0.36 (0.03) & 4 (1) \\
		16 & 0.35 (0.02) & 3 (1) \\
		32 & 0.35 (0.03) & 3 (1)\\
		64 & 0.36 (0.07) & 4 (1) \\
		\bottomrule
	\end{tabular}
%	\vspace*{-3ex}
\end{table}
%\vspace*{-3ex}

\paragraph{Impact of Weighting Distribution} The choice of weighting distribution did not lead to drastic changes in model performance. The $\mathcal{T}$ distribution performs best when weight clipping is used, while $\mathcal{N}$ performs best in the case of gradient penalty. This suggests that the proper choice of distribution is dependent on both the dataset and the Lipschitz regularization used, but the overall framework is robust to reasonable choices. 

We also conducted preliminary experiments using a uniform ($\mathcal{U}$) distribution weighting scheme. Even though the condition $\mathrm{supp}(\mathcal{U}) = \bbR^m$ does not hold for the uniform distribution, we found that this does not adversely affect the performance (see Appendix \ref{sec:addres}). The uniform weighting distribution corresponds to the sinc-kernel in MMD, which is known to be a non-characteristic kernel~\cite{sriperumbudur2010hilbert}. Our results suggest that such kernels could remain effective when used in MMD-GAN, but we did not verify this experimentally. 
%\vspace*{-3ex}
\paragraph{Impact of Number of Random Frequencies} We conducted an experiment to study the impact of the number of random frequencies ($k$) that are sampled from the weighting distribution to compute the ECFD. We ran our best performing model (OCF-GAN-GP) with different values of $k$ from the set $\{1, 4, 8, 16, 32, 64\}$. The FID and KID scores for this experiment are shown in Table \ref{tab:freqexp}. As expected, the score values improve as $k$ increases. However, even for the lowest number of frequencies possible ($k=1$), the performance does not degrade too severely. 
%\hs{put table 1 here. You can have it on the right side of this paragraph.}

\section{Discussion and Conclusion}
 
In this paper, we proposed a novel weighted distance between characteristic functions for training IGMs, and showed that the proposed metric has attractive theoretical properties. We observed experimentally that the proposed model outperforms MMD-GAN and WGAN variants on four benchmark image datasets. Our results indicate that characteristic functions provide an effective alternative means for training IGMs. 

This work opens additional avenues for future research. For example, the empirical CFD used for training may result in high variance gradient estimates (particularly with a small number of sampled frequencies), yet the CFD-trained models attain high performance scores with better convergence in our tests.  The reason for this should be more thoroughly explored. Although we used the gradient penalty proposed by WGAN-GP, there is no reason to constrain the gradient to exactly 1. We believe that an exploration of the geometry of the proposed loss could lead to improvement in the gradient regularizer for the proposed method.

Apart from generative modeling, two sample tests such as MMD have been used for problems such as domain adaptation~\cite{Long2015LearningTF} and domain separation~\cite{Bousmalis2016DomainSN}, among others. The optimized CFD loss function proposed in this work can be used as an alternative loss for these problems.

\paragraph{Acknowledgements} This research is supported by the National Research Foundation Singapore under its AI Singapore Programme (Award Number: AISG-RP-2019-011) to H.~Soh. J.~Scarlett is supported by the Singapore National Research Foundation (NRF) under grant number R-252-000-A74-281.

{\small
\bibliographystyle{ieee_fullname}
\balance
\bibliography{refs}
}
\clearpage
\onecolumn
\appendix
\section{Proofs}

\subsection{Proof of Theorem 1}

% \begin{proof} 
Let $\mathbb{P}_{\mathcal{X}}$ be the data distribution, and let $\mathbb{P}_{g_\theta(\mathcal{Z})}$ be the distribution of $g_{\theta}(\mathbf{z})$ when $\mathbf{z} \sim \mathbb{P}_{\mathcal{Z}}$, with $\mathbb{P}_{\mathcal{Z}}$ being the latent distribution. Recall that the characteristic function of a distribution $\bbQ$ is given by 
\begin{align}
	\varphi_{\bbQ}(\mathbf{t}) = \mathbb{E}_{\mathbf{x}\sim\bbQ}[e^{i\langle \mathbf{t}, \mathbf{x}\rangle}].
\end{align}
The quantity $\mathrm{CFD}_{\omega}^2(\mathbb{P}_{f_\phi({\mathcal{X}})}, \mathbb{P}_{f_\phi(g_\theta(\mathcal{Z}))})$ can then be written as
\begin{align}
    \mathrm{CFD}_{\omega}^2(\mathbb{P}_{f_\phi({\mathcal{X}})}, \mathbb{P}_{f_\phi(g_\theta(\mathcal{Z}))}) = \mathbb{E}_{\mathbf{t}\sim\omega(\mathbf{t};\eta)}\left[\left|\varphi_{\mathcal{X}}(\mathbf{t}) - \varphi_{\theta}(\mathbf{t})\right|^2\right],
\end{align}
where we denote the characteristic functions of $\mathbb{P}_{f_\phi({\mathcal{X}})}$ and $\mathbb{P}_{f_\phi(g_\theta(\mathcal{Z}))}$ by $\varphi_{\mathcal{X}}$ and $\varphi_{\theta}$ respectively, with an implicit dependence of $\phi$.
For notational simplicity, we henceforth denote $\mathrm{CFD}_{\omega}^2(\mathbb{P}_{f_\phi({\mathcal{X}})}, \mathbb{P}_{f_\phi(g_\theta(\mathcal{Z}))})$ by $\mathrm{D}_\psi(\mathbb{P}_{\mathcal{X}},\mathbb{P}_{\theta})$.

% {\bf Proof of continuity.} 
Since the difference of two functions' maximal values is always upper bounded by the maximal gap between the two functions, we have
\begin{align}
    \Big|\underset{\psi\in\Psi}{\sup}\:\mathrm{D}_\psi(\mathbb{P}_{\mathcal{X}},\mathbb{P}_{\theta}) - \underset{\psi\in\Psi}{\sup}\:\mathrm{D}_\psi(\mathbb{P}_{\mathcal{X}},\mathbb{P}_{\theta'})\Big| &\leq \underset{\psi\in\Psi}{\sup}\:|\mathrm{D}_\psi(\mathbb{P}_{\mathcal{X}},\mathbb{P}_{\theta}) - \mathrm{D}_\psi(\mathbb{P}_{\mathcal{X}},\mathbb{P}_{\theta'})|\label{eq:max}\\
    &\le |\mathrm{D}_{\psi^*}(\mathbb{P}_{\mathcal{X}},\mathbb{P}_{\theta}) - \mathrm{D}_{\psi^*}(\mathbb{P}_{\mathcal{X}},\mathbb{P}_{\theta'})| + \epsilon
    \label{eq:optpsi}
\end{align}
where $\psi^* = \{\phi^*, {\eta}^*\}$ denotes any parameters that are within $\epsilon$ of the supremum on the right-hand side of (\ref{eq:optpsi}), and where $\epsilon > 0$ may be arbitrarily small.  Such $\psi^*$ always exists by the definition of supremum.  Subsequently, we define $h_{\theta} = f_{\phi^*} \circ g_\theta$ for compactness.

% Defining $h_{\theta} = f_{\phi^*} \circ g_\theta$ for compactness, 
Let $\omega^*$ denote the distribution $\omega(\mathbf{t})$ associated with $\eta^*$. We further upper bound the right-hand side of \eqref{eq:optpsi} as follows:
 \begin{align}
 	|\mathrm{D}_{\psi^*}(\mathbb{P}_{\mathcal{X}},\mathbb{P}_{\theta}) - \mathrm{D}_{\psi^*}(\mathbb{P}_{\mathcal{X}},\mathbb{P}_{\theta'})| &= \left|\mathbb{E}_{\omega^*(\mathbf{t})}\left[\left|\varphi_{\mathcal{X}}(\mathbf{t}) - \varphi_{\theta}(\mathbf{t})\right|^2\right] - \mathbb{E}_{\omega^*(\mathbf{t})}\left[\left|\varphi_{\mathcal{X}}(\mathbf{t}) - \varphi_{\theta'}(\mathbf{t})\right|^2\right]\right|\\
 	& \overset{(a)}{\leq} \mathbb{E}_{\omega^*(\mathbf{t})}\left[\left|\left|\varphi_{\mathcal{X}}(\mathbf{t}) - \varphi_{\theta}(\mathbf{t})\right|^2 - \left|\varphi_{\mathcal{X}}(\mathbf{t}) - \varphi_{\theta'}(\mathbf{t})\right|^2\right|\right]\label{eq:bound1},
 \end{align}
where $(a)$ uses the linearity of expectation and Jensen's inequality.
 
Since any characteristic function is bounded by $|\varphi_{\bbP}(\mathbf{t})| \leq 1$, the value of $\left|\varphi_{\mathcal{X}}(\mathbf{t}) - \varphi_{\theta}(\mathbf{t})\right|$ for any $\theta$ is upper bounded by 2. Since the function $f(u) = u^2$ is (locally) $4$-Lipschitz over the restricted domain $[0, 2]$, we have
 \begin{align}
 	\left|\left|\varphi_{\mathcal{X}}(\mathbf{t}) - \varphi_{\theta}(\mathbf{t})\right|^2 - \left|\varphi_{\mathcal{X}}(\mathbf{t}) - \varphi_{\theta'}(\mathbf{t})\right|^2\right| &\leq 4\Big|\left|\varphi_{\mathcal{X}}(\mathbf{t}) - \varphi_{\theta}(\mathbf{t})\right| - \left|\varphi_{\mathcal{X}}(\mathbf{t}) - \varphi_{\theta'}(\mathbf{t})\right|\Big| \label{eq:square_removal}\\
 	&\overset{(b)}{\leq} 4\left|\varphi_{\theta}(\mathbf{t}) - \varphi_{\theta'}(\mathbf{t})\right|\\
 	&= 4\left|\mathbb{E}_{\mathbf{z}}\left[e^{i\langle \mathbf{t}, h_{\theta}(\mathbf{z})\rangle}\right] - \mathbb{E}_{\mathbf{z}}\left[e^{i\langle \mathbf{t}, h_{\theta'}(\mathbf{z})\rangle}\right]\right|\\
 	&\overset{(c)}{\leq} 4\mathbb{E}_{\mathbf{z}}\left[\left|e^{i\langle \mathbf{t}, h_{\theta}(\mathbf{z})\rangle} - e^{i\langle \mathbf{t}, h_{\theta'}(\mathbf{z})\rangle}\right|\right],
 	\label{eq:xsqlipschitz}
 \end{align}
where $(b)$ uses the triangle inequality, and $(c)$ uses Jensen's inequality.

%Now, $\left|e^{i\langle \mathbf{t}, h_{\theta}(\mathbf{z})\rangle} - e^{i\langle \mathbf{t}, h_{\theta'}(\mathbf{z})\rangle}\right|$ is the distance between two points lying on a circle of unit radius, centered at origin, in the complex plane. The length of this chord is bounded by the length of the diameter, i.e., 2;  hence, $\left|e^{i\langle \mathbf{t}, h_{\theta}(\mathbf{z})\rangle} - e^{i\langle \mathbf{t}, h_{\theta'}(\mathbf{z})\rangle}\right| \leq 2$. Finally, using the bounded convergence theorem along with inequality (\ref{eq:max}), we obtain
%\begin{align}
%    |\underset{\psi}{\max}\:\mathrm{D}_\psi(\mathbb{P}_{\mathcal{X}},\mathbb{P}_{\theta}) - \underset{\psi}{\max}\:\mathrm{D}_\psi(\mathbb{P}_{\mathcal{X}},\mathbb{P}_{\theta'})| \xrightarrow{\theta \rightarrow \theta'} 0,
%\end{align}
%which proves the continuity statement.

% {\bf Proof of differentiability almost everywhere.} 
In Eq. (\ref{eq:xsqlipschitz}), let $\left|e^{i\langle \mathbf{t}, h_{\theta}(\mathbf{z})\rangle} - e^{i\langle \mathbf{t}, h_{\theta'}(\mathbf{z})\rangle}\right| =: \left|e^{ia} - e^{ib}\right|$, which can be interpreted as the length of the chord that subtends an angle of $|a-b|$ at the center of a unit circle centered at origin. The length of this chord is given by $2\sin \frac{|a-b|}{2}$, and since $2\sin \frac{|a-b|}{2} \leq |a-b|$, we have
 \begin{align}
 \left|e^{i\langle \mathbf{t}, h_{\theta}(\mathbf{z})\rangle} - e^{i\langle \mathbf{t}, h_{\theta'}(\mathbf{z})\rangle}\right| 
    &\leq |\langle \mathbf{t}, h_{\theta}(\mathbf{z})\rangle - \langle \mathbf{t}, h_{\theta'}(\mathbf{z})\rangle|\\
    &\overset{(d)}{\leq} \|\mathbf{t}\| \cdot \|h_{\theta}(\mathbf{z}) - h_{\theta'}(\mathbf{z})\|,
 \end{align}
where $(d)$ uses the Cauchy-Schwarz inequality. 

Furthermore, using the assumption $\sup_{\eta \in \Pi} \mathbb{E}_{\omega(\mathbf{t})}\left[\|\mathbf{t}\|\right] < \infty$, we get
 \begin{align}
     \mathbb{E}_{\omega^*(\mathbf{t})}\left[\mathbb{E}_{\mathbf{z}}\left[\left|e^{i\langle \mathbf{t}, h_{\theta}(\mathbf{z})\rangle} - e^{i\langle \mathbf{t}, h_{\theta'}(\mathbf{z})\rangle}\right|\right]\right] \leq \mathbb{E}_{\omega^*(\mathbf{t})}\left[\|\mathbf{t}\|\right]\mathbb{E}_{\mathbf{z}}\left[\|h_{\theta}(\mathbf{z}) - h_{\theta'}(\mathbf{z})\|\right]
 \end{align}
with the first term being finite.
 
 By assumption, $h$ is locally Lipschitz, i.e., for any pair $(\theta, \mathbf{z})$, there exists a constant $L(\theta, \mathbf{z})$ and an open set $U_{\theta,\mathbf{z}}$ such that $\forall(\theta', \mathbf{z}') \in U_{\theta,\mathbf{z}}$ we have $\|h_{\theta}(\mathbf{z}) - h_{\theta'}(\mathbf{z}')\| \leq L(\theta, \mathbf{z})\|\theta - \theta'\|$. Setting $\mathbf{z}' = \mathbf{z}$ and taking the expectation, we obtain
 \begin{align}
 	\mathbb{E}_{\omega^*(\mathbf{t})}\left[\|\mathbf{t}\|\right]\mathbb{E}_{\mathbf{z}}\left[\|h_{\theta}(\mathbf{z}) - h_{\theta'}(\mathbf{z})\|\right] \leq \mathbb{E}_{\omega^*(\mathbf{t})}\left[\|\mathbf{t}\|\right]\mathbb{E}_{\mathbf{z}}\left[L(\theta, \mathbf{z})\right]\|\theta - \theta'\|\label{eq:bound2}
 \end{align}
 for all $\theta'$ sufficiently close to $\theta$.
 
 Recall also that $\mathbb{E}_{\mathbf{z}}\left[L(\theta, \mathbf{z})\right] < \infty$ by assumption. Combining Eqs. (\ref{eq:bound1}), (\ref{eq:xsqlipschitz}), and (\ref{eq:bound2}), we get 
 \begin{align}
 	|\mathrm{D}_{\psi^*}(\mathbb{P}_{\mathcal{X}},\mathbb{P}_{\theta}) - \mathrm{D}_{\psi^*}(\mathbb{P}_{\mathcal{X}},\mathbb{P}_{\theta'})| \leq 4\mathbb{E}_{\omega^*(\mathbf{t})}\left[\|\mathbf{t}\|\right]\mathbb{E}_{\mathbf{z}}\left[L(\theta, \mathbf{z})\right]\|\theta - \theta'\|,
 \end{align}
 and combining with (\ref{eq:optpsi}) gives
 \begin{align}
 	\Big|\underset{\psi\in\Psi}{\sup}\:\mathrm{D}_\psi(\mathbb{P}_{\mathcal{X}},\mathbb{P}_{\theta}) - \underset{\psi\in\Psi}{\sup}\:\mathrm{D}_\psi(\mathbb{P}_{\mathcal{X}},\mathbb{P}_{\theta'})\Big| &\leq 4\mathbb{E}_{\omega^*(\mathbf{t})}\left[\|\mathbf{t}\|\right]\mathbb{E}_{\mathbf{z}}\left[L(\theta, \mathbf{z})\right]\|\theta - \theta'\| + \epsilon\\
 	&\leq 4\bigg(\sup_{\eta \in \Pi}\mathbb{E}_{\omega(\mathbf{t})} \left[\|\mathbf{t}\|\right]\bigg)\mathbb{E}_{\mathbf{z}}\left[L(\theta, \mathbf{z})\right]\|\theta - \theta'\| + \epsilon.
 \end{align}
Taking the limit $\epsilon \to 0$ on both sides gives
 \begin{align}
 	\Big|\underset{\psi\in\Psi}{\sup}\:\mathrm{D}_\psi(\mathbb{P}_{\mathcal{X}},\mathbb{P}_{\theta}) - \underset{\psi\in\Psi}{\sup}\:\mathrm{D}_\psi(\mathbb{P}_{\mathcal{X}},\mathbb{P}_{\theta'})\Big| \leq 4\bigg(\sup_{\eta \in \Pi}\mathbb{E}_{\omega(\mathbf{t})} \left[\|\mathbf{t}\|\right]\bigg) \mathbb{E}_{\mathbf{z}}\left[L(\theta, \mathbf{z})\right]\|\theta - \theta'\|,
 \end{align}
 which proves that $\underset{\psi\in\Psi}{\sup}\:\mathrm{D}_\psi(\mathbb{P}_{\mathcal{X}},\mathbb{P}_{\theta})$ is locally Lipschitz, and therefore continuous. In addition, Radamacher's theorem~\cite{federer2014geometric} states any locally Lipschitz function is differentiable almost everywhere, which establishes the differentiability claim.
%  that $\underset{\psi}{\max}\:\mathrm{D}_\psi(\mathbb{P}_{\mathcal{X}},\mathbb{P}_{\theta})$ 
 
\subsection{Proof of Theorem 2}

Let $\mathbf{x}_n \sim \mathbb{P}_n$ and $\mathbf{x} \sim \mathbb{P}$.  To study the behavior of $\underset{\psi\in\Psi}{\mathrm{sup}}\:\mathrm{CFD}_{\omega}^2(\mathbb{P}_{n}^{(\phi)},\mathbb{P}^{(\phi)})$, we first consider
 \begin{align}
     % \underset{\psi\in\Psi}{\mathrm{sup}}\:
    \mathrm{CFD}_{\omega}^2(\mathbb{P}_{n}^{(\phi)},\mathbb{P}^{(\phi)}) = \bbE_{\omega(\mathbf{t})}\left[\left|\mathbb{E}_{\mathbf{x}_n}\left[e^{i\langle \mathbf{t}, f_{\phi}(\mathbf{x}_n)\rangle}\right] - \mathbb{E}_{\mathbf{x}}\left[e^{i\langle \mathbf{t}, f_{\phi}(\mathbf{x})\rangle}\right]\right|^2\right]
 \end{align}
 
Since $\left|\mathbb{E}_{\mathbf{x}_n}\left[e^{i\langle \mathbf{t}, f_{\phi}(\mathbf{x}_n)\rangle}\right] - \mathbb{E}_{\mathbf{x}}\left[e^{i\langle \mathbf{t}, f_{\phi}(\mathbf{x})\rangle}\right]\right| \in [0,2]$, using the fact that $u^2 \le 2|u|$ for $u \in [-2,2]$, we have
 \begin{align}
     &\bbE_{\omega(\mathbf{t})}\left[\left|\mathbb{E}_{\mathbf{x}_n}\left[e^{i\langle \mathbf{t}, f_{\phi}(\mathbf{x}_n)\rangle}\right] - \mathbb{E}_{\mathbf{x}}\left[e^{i\langle \mathbf{t}, f_{\phi}(\mathbf{x})\rangle}\right]\right|^2\right] \nonumber \\
    &\qquad \leq 2 \bbE_{\omega(\mathbf{t})}\left[\left|\mathbb{E}_{\mathbf{x}_n,\mathbf{x}}\left[e^{i\langle \mathbf{t}, f_{\phi}(\mathbf{x}_n)\rangle} - e^{i\langle \mathbf{t}, f_{\phi}(\mathbf{x})\rangle}\right]\right|\right]\\
     &\qquad\overset{(a)}{\leq} 2 \bbE_{\omega(\mathbf{t})}\left[\mathbb{E}_{\mathbf{x}_n,\mathbf{x}}\left[\left|e^{i\langle \mathbf{t}, f_{\phi}(\mathbf{x}_n)\rangle} - e^{i\langle \mathbf{t}, f_{\phi}(\mathbf{x})\rangle}\right|\right]\right]\\
     &\qquad\overset{(b)}{\leq} 2 \bbE_{\omega(\mathbf{t})}\left[\mathbb{E}_{\mathbf{x}_n,\mathbf{x}}\left[ \min\left\{2, \left|\langle \mathbf{t}, f_{\phi}(\mathbf{x}_n)\rangle - \langle \mathbf{t}, f_{\phi}(\mathbf{x})\rangle\right| \right\}\right]\right] \\
     &\qquad\overset{(c)}{\leq} 2 \bbE_{\omega(\mathbf{t})}\left[\mathbb{E}_{\mathbf{x}_n,\mathbf{x}}\left[ \min\left\{2, \|\mathbf{t}\| \cdot \|f_{\phi}(\mathbf{x}_n) - f_{\phi}(\mathbf{x})\| \right\}\right]\right], \label{eq:post_cauchy}
 \end{align}
where $(a)$ uses Jensen's inequality, $(b)$ uses the geometric properties stated following Eq.~\eqref{eq:xsqlipschitz} and the fact that $|e^{ia} - e^{ib}| \le 2$, and $(c)$ uses the Cauchy-Schwarz inequality.

For brevity, let $T_{\max} = \sup_{\eta \in \Pi}\bbE_{\omega(\mathbf{t})}\left[\|\mathbf{t}\|\right]$, which is finite by assumption. Interchanging the order of the expectations in Eq.~\eqref{eq:post_cauchy} and applying Jensen's inequality (to $\mathbb{E}_{\omega(\mathbf{t})}$ alone) and the concavity of $f(u) = \min\{2,u\}$, we can continue the preceding upper bound as follows:
\begin{align}
     &\bbE_{\omega(\mathbf{t})}\left[\left|\mathbb{E}_{\mathbf{x}_n}\left[e^{i\langle \mathbf{t}, f_{\phi}(\mathbf{x}_n)\rangle}\right] - \mathbb{E}_{\mathbf{x}}\left[e^{i\langle \mathbf{t}, f_{\phi}(\mathbf{x})\rangle}\right]\right|^2\right] \nonumber \\
     &\qquad\leq 2 \mathbb{E}_{\mathbf{x}_n,\mathbf{x}}\left[ \min\left\{2, T_{\rm max} \|f_{\phi}(\mathbf{x}_n) - f_{\phi}(\mathbf{x})\|  \right\}\right]\\
     &\qquad\overset{(d)}{\leq}  2 \mathbb{E}_{\mathbf{x}_n,\mathbf{x}}\left[ \min\left\{2, T_{\rm max} L_f \|\mathbf{x}_n - \mathbf{x}\|  \right\}\right], \label{eq:weak_final}
      % &\qquad\overset{(d)}{\leq} 2 \bbE_{\omega(\mathbf{t})}\left[\|\mathbf{t}\|\right]\mathbb{E}\left[\|f_{\phi}(\mathbf{x}_n) - f_{\phi}(\mathbf{x})\|\right]
      % &\qquad\overset{(e)}{\leq} 2 \bbE_{\omega(\mathbf{t})}\left[\|\mathbf{t}\|\right]L_f\mathbb{E}\left[\|\mathbf{x}_n - \mathbf{x}\|\right], \label{eq:weak_final}
\end{align}
where $(d)$ defines $L_f$ to be the Lipschitz constant of $f_\phi$, with is independent of $\phi$ by assumption. %, i.e., $\max_{\phi}\|f_\phi\|_{\mathrm{Lip}} \leq L_f < \infty$. 

% From Assumption 2, we have $\sup_{\eta \in \Pi}\bbE_{\omega(\mathbf{t})}\left[\|\mathbf{t}\|\right] < \infty$, and the assumption $\bbP_n \xrightarrow{D} \bbP$ implies $\mathbb{E}\left[\|\mathbf{x}_n - \mathbf{x}\|\right] \to 0$.  Substituting into \eqref{eq:weak_final} yields $\underset{\psi\in\Psi}{\mathrm{sup}}\:\mathrm{CFD}_{\omega}^2(\mathbb{P}_{n},\mathbb{P}) \rightarrow 0$, as required.

Observe that $g(u) = \min\{2,T_{\max}L_f |u|\}$ is a bounded Lipschitz function of $u$.  By the Portmanteau theorem (\cite{klenke2013probability}, Thm. 13.16), convergence in distribution $\bbP_n \xrightarrow{D} \bbP$ implies that $\mathbb{E}[ g( \|\mathbf{x}_n - \mathbf{x}\| ) ] \to 0$ for any such $g$, and hence \eqref{eq:weak_final} yields $\underset{\psi\in\Psi}{\mathrm{sup}}\:\mathrm{CFD}_{\omega}^2(\mathbb{P}_{n}^{(\phi)},\mathbb{P}^{(\phi)}) \rightarrow 0$ (upon taking $\sup_{\psi\in\Psi}$ on both sides), as required.

% \af{Only if proof}
% \end{proof}

\subsection{Discussion on an ``only if'' Counterpart to Theorem 2}

Theorem 2 shows that, under some technical assumptions, the function
$\underset{\psi \in \Psi}{\mathrm{sup}}\:\mathrm{CFD}_{\omega}^2(\mathbb{P}_{n}^{(\phi)},\mathbb{P}^{(\phi)})$ 
satisfies continuity in the weak toplogy, i.e., 
    $$\mathbb{P}_n \stackrel{{D}}{\to} \mathbb{P} \implies \underset{\psi \in \Psi}{\mathrm{sup}}\:\mathrm{CFD}_{\omega}^2(\mathbb{P}_{n}^{(\phi)},\mathbb{P}^{(\phi)}) \to 0. $$ 
where $\mathbb{P}_n \stackrel{{D}}{\to} \mathbb{P}$ denotes convergence in distribution.

Here we discuss whether the opposite is true: Does $ \underset{\psi \in \Psi}{\mathrm{sup}}\:\mathrm{CFD}_{\omega}^2(\mathbb{P}_{n}^{(\phi)},\mathbb{P}^{(\phi)}) \to 0$ imply that $\mathbb{P}_n \stackrel{{D}}{\to} \mathbb{P}$?
In general, the answer is negative.  For example:
\begin{itemize}
    \item If $\Phi$ only contains the function $\phi(x) = 0$, then $\mathbb{P}^{(\phi)}$ is always the distribution corresponding to deterministically equaling zero, so {\em any} two distributions give zero CFD.
    \item If $\omega(t)$ has bounded support, then two distributions $\mathbb{P}_1,\mathbb{P}_2$ whose characteristic functions only differ for $t$ values outside that support may still give $\mathbb{E}_{\omega(t)} \big[ |\varphi_{\mathbb{P}_1}(t) - \varphi_{\mathbb{P}_2}(t) |^2 \big] = 0$. 
\end{itemize}
In the following, however, we argue that the answer is positive when $\{f_{\phi}\}_{\phi \in \Phi}$ is ``sufficiently rich'' and $\{\omega\}_{\eta \in \Pi}$ is ``sufficiently well-behaved''.

Rather than seeking the most general assumptions that formalize these requirements, we focus on a simple special case that still captures the key insights, assuming the following:
\begin{itemize}
    \item There exists $L > 0$ such that $\{f_{\phi}\}_{\phi \in \Phi}$ includes all {\em linear} functions that are $L$-Lipschitz;
    \item There exists $\eta \in \Pi$ such that $\omega(\mathbf{t})$ has support $\mathbb{R}^m$, where $m$ is the output dimension of $f_\phi$.
\end{itemize}
To give examples of these, note that neural networks with ReLU activations can implement arbitrary linear functions (with the Lipschitz condition amounting to bounding the weights), and note that the second assumption is satisfied by any Gaussian $\omega(\mathbf{t})$ with a fixed positive-definite covariance matrix.

In the following, let $\mathbf{x}_n \sim \mathbb{P}_n$ and $\mathbf{x} \sim \mathbb{P}^{(\phi)}$.  We will prove the contrapositive statement:
    $$ \bbP_n \stackrel{{D}}{\not\to} \mathbb{P}^{(\phi)} \implies \underset{\psi \in \Psi}{\mathrm{sup}}\:\mathrm{CFD}_{\omega}^2(\mathbb{P}_{n},\mathbb{P}^{(\phi)}) \not\to 0. $$
By the Cram\'er-Wold theorem~\cite{cramer1936some}, $\bbP_n \stackrel{{D}}{\not\to} \mathbb{P}^{(\phi)}$ implies that we can find constants $c_1,\dotsc,c_d$ such that
\begin{equation}
    \sum_{i=1}^d c_i \mathbf{x}_n^{(i)}  \stackrel{{D}}{\not\to}  \sum_{i=1}^d c_i \mathbf{x}^{(i)}, \label{eq:non_conv}
\end{equation}
where $\mathbf{x}^{(i)},\mathbf{x}_n^{(i)}$ denote the $i$-th entries of $\mathbf{x},\mathbf{x}_n$, with $d$ being their dimension.

Recall that we assume $\{f_{\phi}\}_{\phi \in \Phi}$ includes all linear functions from $\mathbb{R}^d$ to $\mathbb{R}^m$ with Lipschitz constant at most $L > 0$.   Hence, we can select $\phi \in \Phi$ such that every entry of $f_{\phi}(x)$ equals $\frac{1}{Z} \sum_{i=1}^d c_i x^{(i)}$, where $Z$ is sufficiently large so that the Lipschitz constant of this $f_{\phi}$ is at most $L$.  However, for this $\phi$, \eqref{eq:non_conv} implies that $f_{\phi}(\mathbf{x}_n) \stackrel{{D}}{\not\to} f_{\phi}(\mathbf{x})$, which in turn implies that $|\varphi_{\mathbb{P}_n^{(\phi)}}(t) - \varphi_{\mathbb{P}^{(\phi)}}(t)|$ is bounded away from zero for all $t$ in some set $\mathcal{T}$ of positive Lebesgue measure.

Choosing $\omega(\mathbf{t})$ to have support $\mathbb{R}^m$ in accordance with the second technical assumption above, it follows that $\mathbb{E}_{\omega(t)} \big[ |\varphi_{\mathbb{P}_1^{(\phi)}}(t) - \varphi_{\mathbb{P}_2^{(\phi)}}(t) |^2 \big] \not\to 0$ and hence $\underset{\psi \in \Psi}{\mathrm{sup}}\:\mathrm{CFD}_{\omega}^2(\mathbb{P}_{n}^{(\phi)},\mathbb{P}^{(\phi)}) \not\to 0$.

\section{Implementation Details}

\begin{figure}
	\centering
	\subfloat[$\mathcal{D}_1$]{\includegraphics[width=0.48\linewidth]{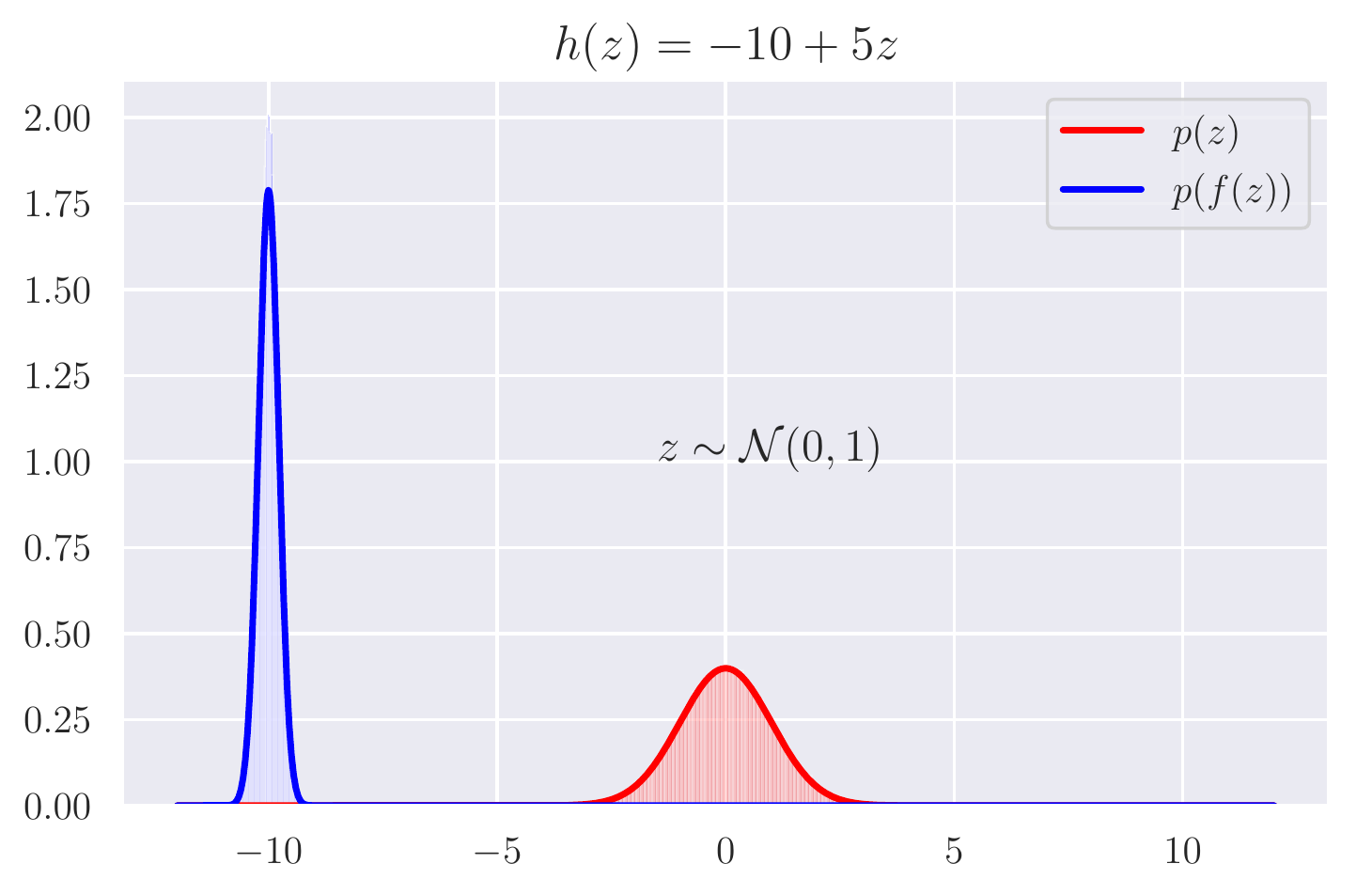}
		\label{fig:uni-pdf}}
	\subfloat[$\mathcal{D}_2$]{\includegraphics[width=0.48\linewidth]{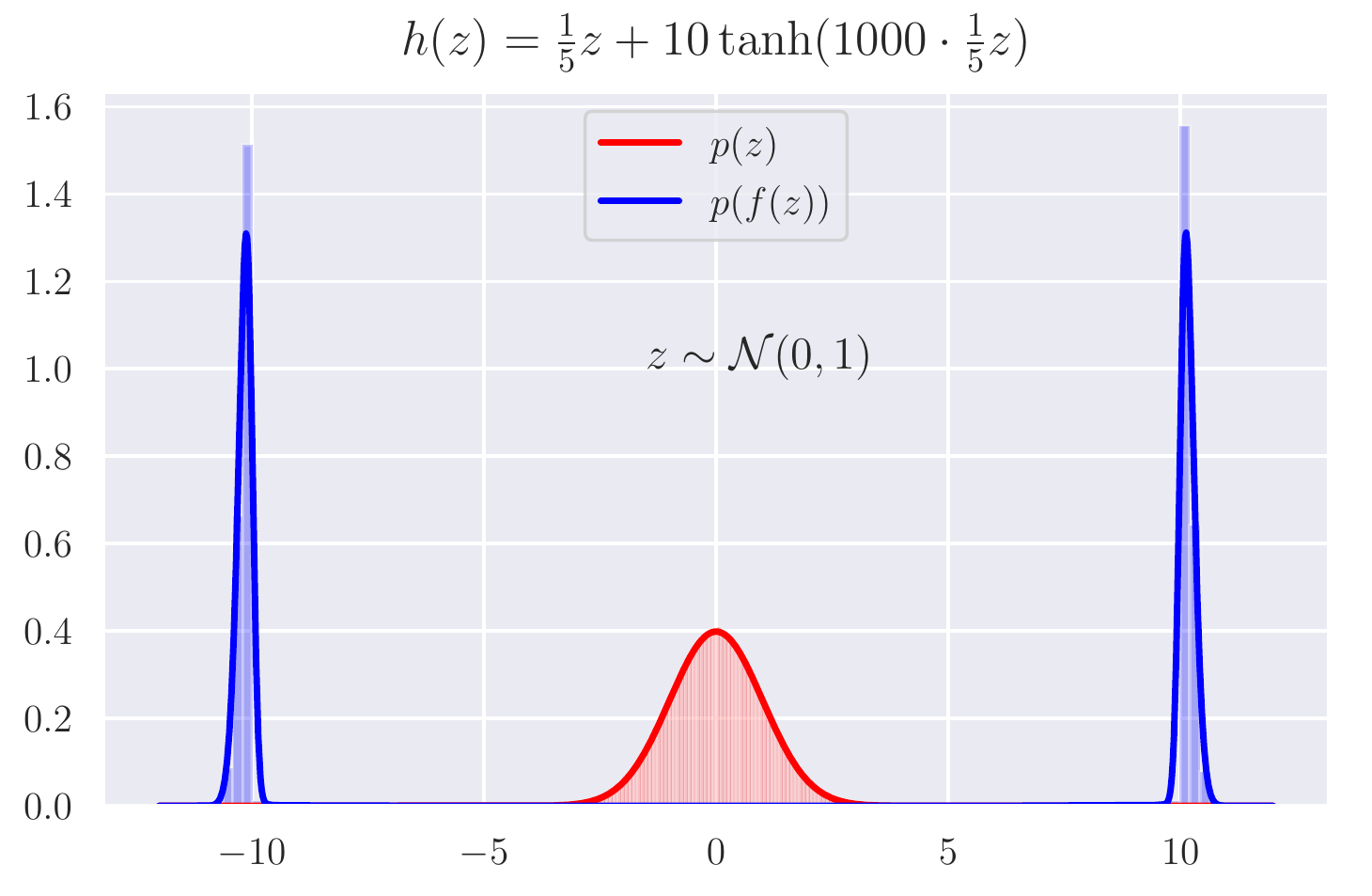}
		\label{fig:bi-pdf}}
	\caption{The PDFs of $\mathcal{D}_1$ and $\mathcal{D}_2$ (in blue) estimated using Kernel Density Estimation (KDE) along with the true distribution $p(z)$ (in red).}
\end{figure}
\subsection{Synthetic Data Experiments}
\label{sec:synthetic}
The synthetic data was generated by first sampling $z \sim \calN(0,1)$ and then applying a function $h$ to the samples. We constructed distributions of two types: a scale-shift unimodal distribution $\mathcal{D}_1$ and a ``scale-split-shift" bimodal distribution $\mathcal{D}_2$. The function $h$ for the two distributions are defined as follows:
\begin{itemize}
	\item $\mathcal{D}_1$: $h(z) = \mu + \sigma z$; we set $\mu=-10$ and $\sigma=\frac{1}{5}$. This shifts the mean of the distribution to $-10$, resulting in the $\calN(-10, \frac{1}{5^2})$ distribution. Fig.~\ref{fig:uni-pdf} shows the PDF (and histogram) of the original distribution $p(z)$ and the distribution of $h(z)$, which is approximated using Kernel Density Estimation (KDE).
	\item $\mathcal{D}_2$: $h(z) = \alpha z + \beta\tanh(\gamma\alpha z)$; we set $\alpha=\frac{1}{5}$, $\beta=10$, $\gamma=100$. This splits the distribution into two modes and shifts the two modes to $-10$ and $+10$. Fig.~\ref{fig:bi-pdf} shows the PDF (and histogram) of the original distribution $p(z)$ and the distribution of $h(z)$,  which is approximated using KDE.
\end{itemize}

For the two cases described above, there are two transformation functions that will lead to the same distribution. In each case, the second transformation function is given by:
\begin{itemize}
	\item $\mathcal{D}_1$: $g(z) = \mu - \sigma z$
	\item $\mathcal{D}_2$: $g(z) = -\alpha z + \beta\tanh(-\gamma\alpha z)$
\end{itemize}

As there are two possible correct transformation functions ($h$ and $g$) that the GANs can learn, we computed the Mean Absolute Error (MAE) as follows
\begin{align}
\mathrm{MAE} = \min\left(\mathbb{E}_{z}\left[|h(z) - \hat{h}(z)|\right], \mathbb{E}_{z}\left[|g(z) - \hat{h}(z)|\right]\right),\label{eq:mae}
\end{align}
where $\hat{h}$ is the transformation learned by the generator. We estimated the expectations in Eq.~(\ref{eq:mae}) using 5000 samples.

For the generator and critic network architectures, we followed \cite{Zaheer2018ConnoisseurC}. Specifically, the generator is a multi-layer perceptron (MLP) with 3 hidden layers of sizes 7, 13, 7, and the Exponential linear unit (ELU) non-linearity between the layers. The critic network is also an MLP with 3 hidden layers of sizes 11, 29, 11, and the ELU non-linearity between the layers. The inputs and outputs of both networks are one-dimensional. We used the RMSProp optimizer with a learning rate of 0.001 for all models. The batch size was set to 50, and 5 critic updates were performed per generator iteration. We trained the models for 10000 and 20000 generator iterations for $\mathcal{D}_1$ and $\mathcal{D}_2$ respectively. For all the models that rely on weight clipping, clipping in the range $[-0.01,0.01]$ for $\mathcal{D}_2$ resulted in poor performance, so we modified the range to $[-0.1,0.1]$.

We used a mixture of 5 RBF kernels for MMD-GAN~\cite{Li2017MMDGT}, and a mixture of 5 RQ kernels and gradient penalty (as defined in \cite{Binkowski2018DemystifyingMG}) for MMD-GAN-GP. For the CF-GAN variants, we used a single weighting distribution (Student-t and Gaussian for $\mathcal{D}_1$ and $\mathcal{D}_2$ respectively). The gradient penalty trade-off parameter ($\lambda_{\mathrm{GP}}$) for WGAN-GP was set to 1 for $\mathcal{D}_1$ as the value of 10 led to erratic performance.

\subsection{Image Generation}
\label{sec:imagexp}

\paragraph{CF-GAN} Following \cite{Li2017MMDGT}, a decoder was also connected to the critic in CF-GAN to reconstruct the input to the critic. This encourages the critic to learn a representation that has a high mutual information with the input. The auto-encoding objective is optimized along with the discriminator, and the final objective is given by
\begin{align}
\underset{\theta}{\inf}\:\underset{\psi}{\sup}\:&\mathrm{CFD}_{\omega}^2(\mathbb{P}_{f_\phi({\mathcal{X}})}, \mathbb{P}_{f_\phi(g_\theta(\mathcal{Z}))})- \lambda_1\mathbb{E}_{\textbf{u} \in \mathcal{X}\cup g_\theta(\mathcal{Z})}\left[\mathcal{D}(\textbf{u}, f_\phi^d(f_\phi(\textbf{u})))\right],
\end{align}
where $f_\phi^d$ is the decoder network, $\lambda_1$ is the regularization parameter, and $\mathcal{D}$ is the discrepancy between the two data-points (e.g., squared error, cross-entropy, etc.). Although the decoder is interesting from an auto-encoding perspective of the representation learned by $f_\phi$, we found that the removal of the decoder did not impact the performance of the model; this can be seen by the results of OCF-GAN-GP, which does not use a decoder network.

We also reduced the feasible set~\cite{Li2017MMDGT} of $f_\phi$, which amounts to an additive penalty of $\lambda_2 \min\left(\bbE[f_\phi(\textbf{x})] - \bbE[f_\phi(g_\theta(\textbf{z}))], 0\right)$. We observed in our experiments that this led to improved stability of training, especially for the models that use weight clipping to enforce Lipschitz condition. For more details, we refer the reader to \cite{Li2017MMDGT}. 

\paragraph{Network and Hyperparameter Details} We used DCGAN-like generator $g_\theta$ and critic $f_\phi$ architectures, same as \cite{Li2017MMDGT} for all models. Specifically, both $g_\theta$ and $d_\phi$ are fully convolutional networks with the following structures:
\begin{itemize}
	\item $g_\theta$: upconv(256) $\to$ bn $\to$ relu $\to$ upconv(128) $\to$ bn $\to$ relu $\to$ upconv(64) $\to$ bn $\to$ relu $\to$ upconv($c$) $\to$ tanh;
	\item $f_\phi$: conv(64) $\to$ leaky-relu(0.2) $\to$ conv(128) $\to$ bn $\to$ leaky-relu(0.2) $\to$ conv(256) $\to$ bn $\to$ leaky-relu(0.2) $\to$ conv($m$),
\end{itemize}
where conv, upconv, bn, relu, leaky-relu, and tanh refer to convolution, up-convolution, batch-normalization, ReLU, LeakyReLU, and Tanh layers respectively. The decoder $f_{\phi^d}$ (whenever used) is also a DCGAN-like decoder. The generator takes a $k$-dimensional Gaussian latent vector as the input and outputs a $32\times 32$ image with $c$ channels. The value of $k$ was set differently depending on the dataset: MNIST (10), CIFAR10 (32), STL10 (32), and CelebA (64). The output dimensionality of the critic network ($m$) was set to 10 (MNIST) and 32 (CIFAR10, STL10, CelebA) for the MMD-GAN and CF-GAN models and 1 for WGAN and WGAN-GP. The batch normalization layers in the critic were omitted for WGAN-GP and OCF-GAN-GP (as suggested by \cite{Gulrajani2017ImprovedTO}).
  
  RMSProp optimizer was used with a learning rate of $5 \times 10^{-5}$. All models were optimized with a batch size of 64 for 125000 generator iterations (50000 for MNIST)  with 5 critic updates per generator iteration. We tested CF-GAN variants with two weighting distributions: Gaussian ($\mathcal{N}$) and Student-t ($\mathcal{T}$) (with 2 degrees of freedom). We also conducted preliminary experiments using Laplace ($\mathcal{L}$) and Uniform ($\mathcal{U}$) weighting distributions (see Table \ref{tab:cifar-full}). For CF-GAN, we tested with 3 scale parameters for $\mathcal{N}$ and $\mathcal{T}$ from the set $\{0.2, 0.5, 1\}$, and we report the best results. The trade-off parameter for the auto-encoder penalty ($\lambda_1$) and feasible-set penalty ($\lambda_2$) were set to 8 and 16 respectively, as in \cite{Li2017MMDGT}. For OCF-GAN-GP, the trade-off for the gradient penalty was set to 10, same as WGAN-GP. The number of random frequencies $k$ used for computing ECFD for all CF-GAN models was set to 8. For MMD-GAN, we used a mixture of five RBF kernels $k_\sigma(x, x') = \exp\left(\frac{\|x-x'\|^2}{2\sigma^2}\right)$ with different scales ($\sigma$) in $\Sigma = \{1, 2, 4, 8, 16\}$ as in \cite{Li2017MMDGT}. For MMD-GAN-GP$_{L2}$, we used a mixture of rational quadratic kernels $k_\sigma(x, x') = \left(1+\frac{\|x-x'\|^2}{2\alpha}\right)^{-\alpha}$ with $\alpha$ in $\mathcal{A} = \{0.2, 0.5, 1, 2, 5\}$; the trade-off parameters of the gradient and L2 penalties were set according to \cite{Binkowski2018DemystifyingMG}. 

\paragraph{Evaluation Metrics} 
We compared the different models using three evaluation metrics: Fr\'echet Inception Distance (FID)~\cite{salimans2016improved}, Kernel Inception Distance (KID)~\cite{Binkowski2018DemystifyingMG}, and Precision-Recall (PR) for Generative models~\cite{sajjadi2018assessing}. All evaluation metrics use features extracted from the \texttt{pool3} layer (2048 dimensional) of an Inception network pre-trained on ImageNet, except for MNIST, for which we used a LeNet5 as the feature extractor. FID fits Gaussian distributions to Inception features of the real and fake images and then computes the Fr\'echet distance between the two Gaussians. On the other hand, KID computes the MMD between the Inception features of the two distributions using a polynomial kernel of degree 3. This is equivalent to comparing the first three moments of the two distributions.

Let $\{x^r_i\}_{i=1}^n$ be samples from the data distribution $\bbP_r$ and $\{x^g_i\}_{i=1}^m$ be samples from the GAN generator distribution $\bbQ_\theta$. Let $\{z^r_i\}_{i=1}^n$ and $\{z^g_i\}_{i=1}^m$ be the feature vectors extracted from the Inception network for $\{x^r_i\}_{i=1}^n$ and $\{x^g_i\}_{i=1}^m$ respectively. The FID and KID are then given by
\begin{align}
\mathrm{FID}(\mathbb{P}_r, \mathbb{Q}_\theta) =& ||\mu_r - \mu_g||^2 + \mathrm{Tr} (\Sigma_r + \Sigma_g - 2 (\Sigma_r \Sigma_g)^{1/2}),\label{eq:fid}\\
\mathrm{KID}(\mathbb{P}_r, \mathbb{Q}_\theta) =& \frac{1}{n(n-1)}\sum_{i=1}^n\sum_{j=1,j\neq i}^n\left[\kappa(z^r_i,z^r_j)\right]\nonumber\\
&+ \frac{1}{m(m-1)}\sum_{i=1}^m\sum_{j=1,j\neq i}^m\left[\kappa(z^g_i,z^g_j)\right]\label{eq:kid}\\
&-\frac{2}{mn}\sum_{i=1}^n\sum_{j=1}^m\left[\kappa(z^r_i,z^g_j)\right], \nonumber
\end{align}
where ($\mu_r$, $\Sigma_r$) and ($\mu_g$, $\Sigma_g$) are the sample mean \& covariance matrix of the inception features of the real and generated data distributions, and $\kappa$ is a polynomial kernel of degree 3, i.e.,
\begin{align}
\kappa(x, y) = \left(\frac{1}{m}\langle x, y \rangle + 1\right)^3,
\end{align}
where $m$ is the dimensionality of the feature vectors.

Both FID and KID give single-value scores, and PR gives a two-dimensional score which disentangles the quality of generated samples from the coverage of the data distribution. For more details about PR, we refer the reader to \cite{sajjadi2018assessing}. In brief, PR is defined by a pair $F_8$ (recall) and $F_{1/8}$ (precision), which represent the coverage and sample quality respectively~\cite{sajjadi2018assessing}.

We used 50000 (10000 for PR) random samples from the different GANs to compute the FID and KID scores. For MNIST and CIFAR10, we compared against the standard test sets, while for CelebA and STL10, we compared against 50000 random images sampled from the dataset. Following \cite{Binkowski2018DemystifyingMG}, we computed FID using 10 bootstrap resamplings and KID by sampling 1000 elements (without replacement) 100 times.

\section{Additional Results}
\label{sec:addres}

\begin{figure}
	\centering
	\includegraphics[width=0.6\linewidth]{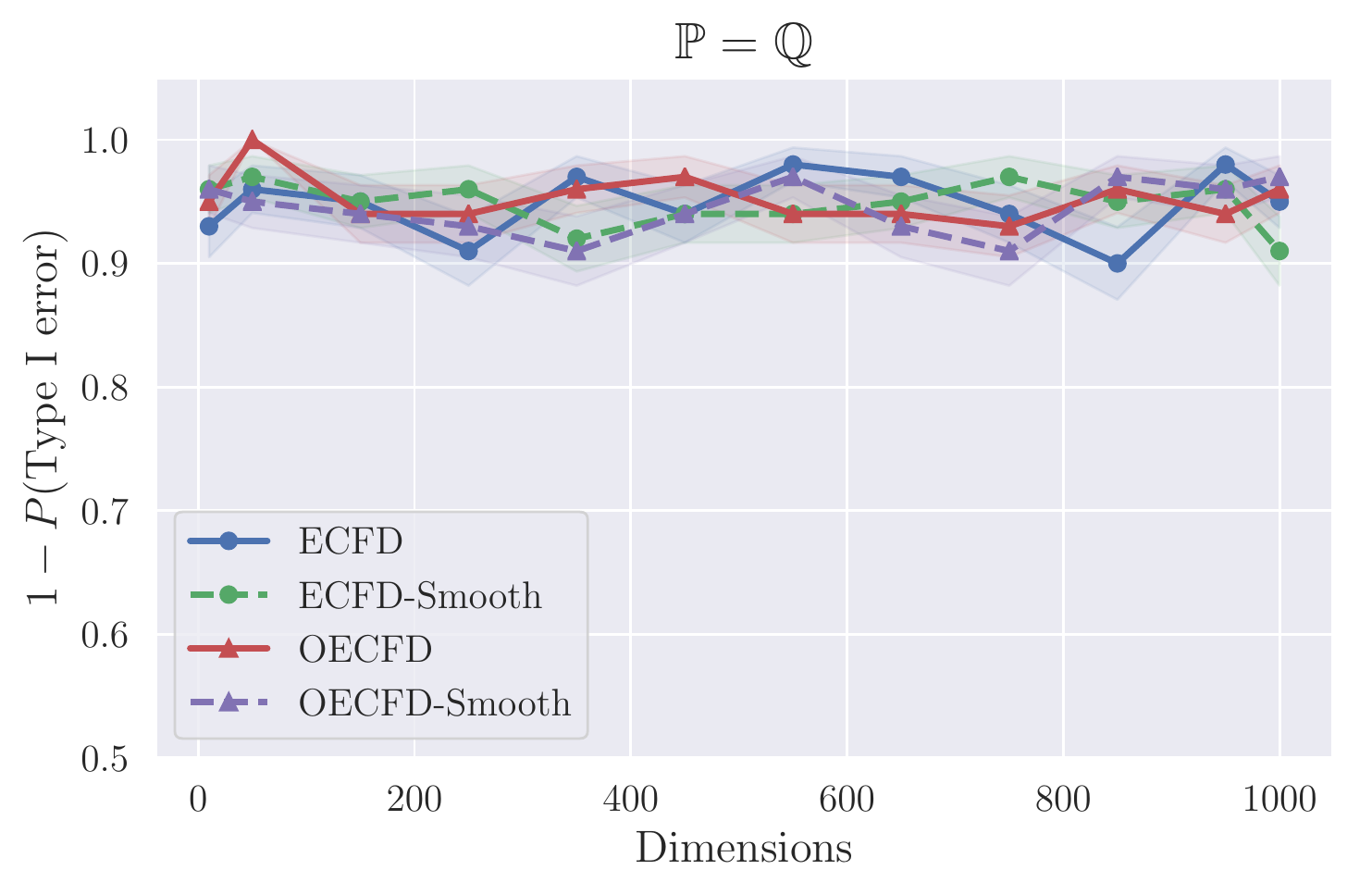}
	\caption{Probability of correctly accepting the null hypothesis $\bbP = \bbQ$ for various numbers of dimensions and different variants of ECFD.}
	\label{fig:toyexp1same}
\end{figure}
Fig.~\ref{fig:toyexp1same} shows the probability of accepting the null hypothesis $\bbP = \bbQ$ when it is indeed correct for different two sample tests based on ECFs. As mentioned in the main text, the optimization of the parameters of the weighting distribution does not hamper the ability of the test to correctly recognize the cases that $\bbP = \bbQ$.

Table \ref{tab:cifar-full} shows the FID and KID scores for various models for the CIFAR10, STL10, and CelebA datasets, including results for the smoothed version of ECFD and Laplace ($\mathcal{L}$) \& Uniform ($\mathcal{U}$) weighting distributions. The FID and KID scores for the MNIST dataset are shown in Table \ref{tab:mnist}. 

\begin{figure*}
	\centering
	\subfloat[WGAN]{\includegraphics[width=0.4\linewidth]{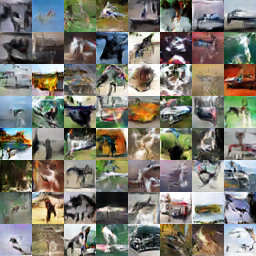}}\quad
	\subfloat[WGAN-GP]{\includegraphics[width=0.4\linewidth]{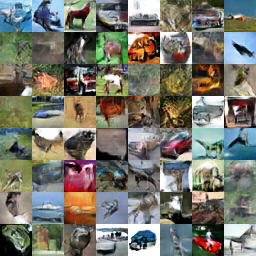}}\quad\\
	\subfloat[MMD-GAN]{\includegraphics[width=0.4\linewidth]{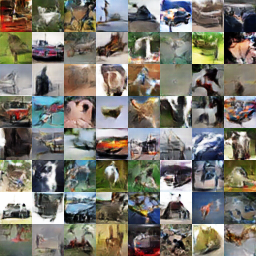}}\quad
	\subfloat[MMD-GAN-GP]{\includegraphics[width=0.4\linewidth]{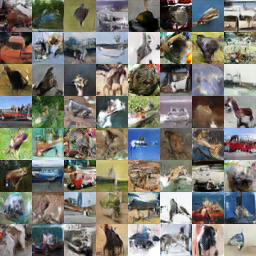}}\quad\\
	\subfloat[OCF-GAN-GP]{\includegraphics[width=0.4\linewidth]{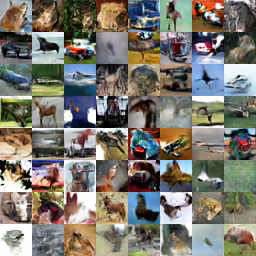}}\quad
	\subfloat[CIFAR10 Test Set]{\includegraphics[width=0.4\linewidth]{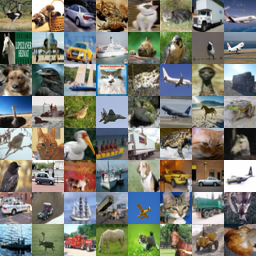}}\\
	\caption{Image samples from the different models for the CIFAR10 dataset.}
	\label{fig:samples-cifar10}
\end{figure*}

\begin{figure*}
	\centering
	\subfloat[WGAN]{\includegraphics[width=0.4\linewidth]{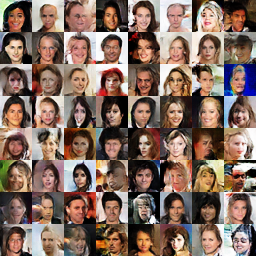}}\quad
	\subfloat[WGAN-GP]{\includegraphics[width=0.4\linewidth]{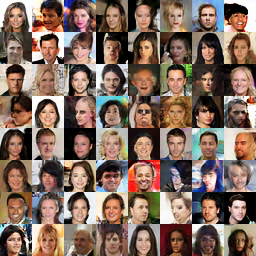}}\quad\\
	\subfloat[MMD-GAN]{\includegraphics[width=0.4\linewidth]{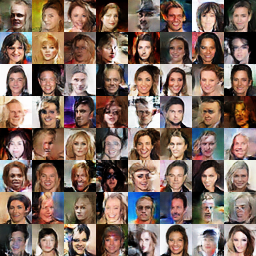}}\quad
	\subfloat[MMD-GAN-GP]{\includegraphics[width=0.4\linewidth]{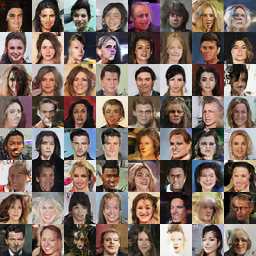}}\quad\\
	\subfloat[OCF-GAN-GP]{\includegraphics[width=0.4\linewidth]{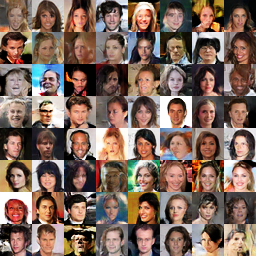}}\quad
	\subfloat[CelebA Real Samples]{\includegraphics[width=0.4\linewidth]{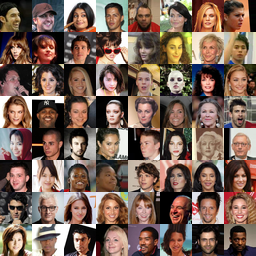}}\\
	\caption{Image samples from the different models for the CelebA dataset.}
	\label{fig:samples-celeba}
\end{figure*}

\begin{figure*}
	\centering
	\subfloat[WGAN]{\includegraphics[width=0.4\linewidth]{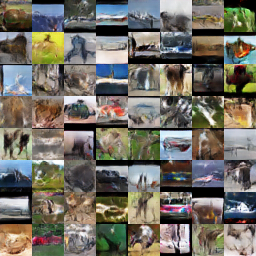}}\quad
	\subfloat[WGAN-GP]{\includegraphics[width=0.4\linewidth]{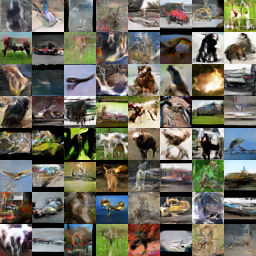}}\quad\\
	\subfloat[MMD-GAN]{\includegraphics[width=0.4\linewidth]{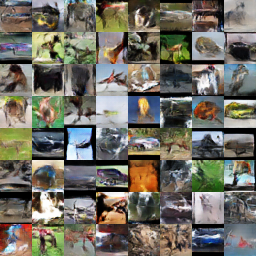}}\quad
	\subfloat[MMD-GAN-GP]{\includegraphics[width=0.4\linewidth]{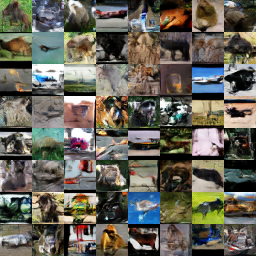}}\quad\\
	\subfloat[OCF-GAN-GP]{\includegraphics[width=0.4\linewidth]{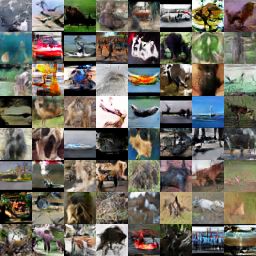}}\quad
	\subfloat[STL10 Test Set]{\includegraphics[width=0.4\linewidth]{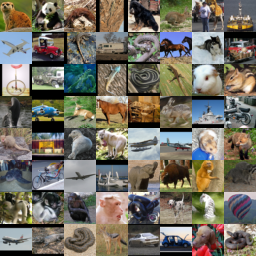}}\\
	\caption{Image samples from the different models for the STL10 dataset.}
	\label{fig:samples-stl10}
\end{figure*}

\begin{figure}
	\centering
	\subfloat[$k$ = 1]{\includegraphics[width=0.22\columnwidth]{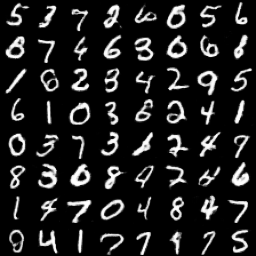}}\quad
	\subfloat[$k$ = 4]{\includegraphics[width=0.22\columnwidth]{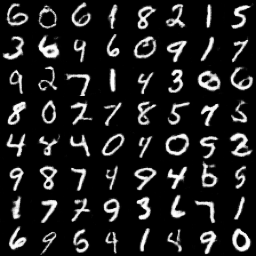}}\\
	\subfloat[$k$ = 8]{\includegraphics[width=0.22\columnwidth]{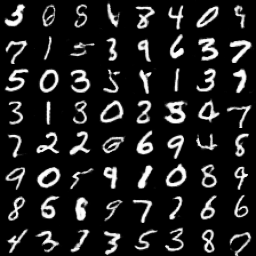}}\quad
	\subfloat[$k$ = 16]{\includegraphics[width=0.22\columnwidth]{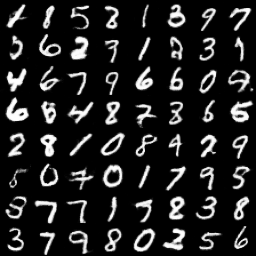}}\\
	\subfloat[$k$ = 32]{\includegraphics[width=0.22\columnwidth]{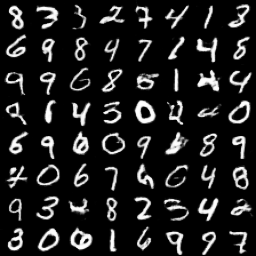}}\quad
	\subfloat[$k$ = 64]{\includegraphics[width=0.22\columnwidth]{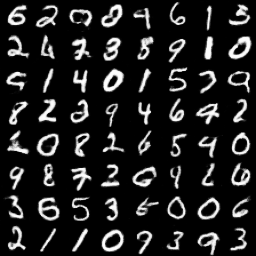}}\\
	\caption{Image samples from OCF-GAN-GP for the MNIST dataset trained using different numbers of random frequencies ($k$).}
	\label{fig:samples-mnist-freq}
\end{figure}

\begin{figure}
	\centering
	\subfloat{\includegraphics[width=0.95\columnwidth]{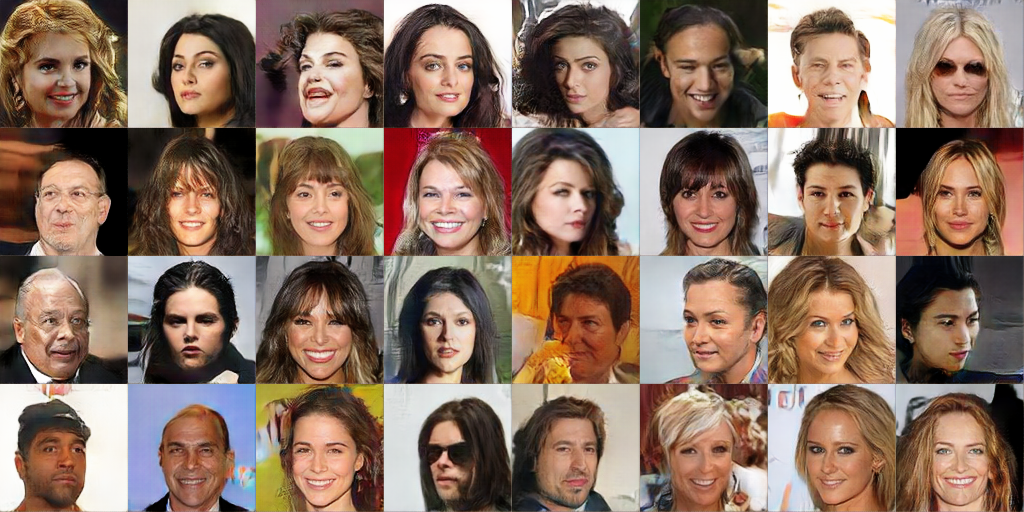}}
	\caption{Image samples for the $128 \times 128$ CelebA dataset generated by OCF-GAN-GP with a ResNet generator.}
	\label{fig:celebsamples}
%    \vspace*{-2ex}
\end{figure}

Figures \ref{fig:samples-cifar10}, \ref{fig:samples-celeba}, and \ref{fig:samples-stl10} show random images generated by different GAN models for CIFAR10, CelebA, and STL10 datasets respectively. The images generated by models that do not use gradient penalty (WGAN and MMD-GAN) are less sharp and have more artifacts compared to their GP counterparts. Fig.~\ref{fig:samples-mnist-freq} shows random images generated from OCF-GAN-GP($\mathcal{N}$) trained on the MNIST dataset with a different number of random frequencies ($k$). It is interesting to note that the change in sample quality is imperceptible even when $k=1$. Figure \ref{fig:celebsamples} shows additional samples from OCF-GAN-GP with a ResNet generator trained on CelebA $128 \times 128$.  

\begin{table}
	\caption{FID and KID ($\times 10^3$) scores (lower is better) for CIFAR10, STL10, and CelebA datasets. Results are averaged over 5 random runs wherever the standard deviation is indicated in parentheses.}
	\label{tab:cifar-full}
	\centering
	\small 
	\begin{tabular}{llrrrrrr} 
		\toprule
		\multirow{2}{*}{Model}                        & \multirow{2}{*}{Kernel/}  & \multicolumn{2}{c}{CIFAR10} & \multicolumn{2}{c}{STL10} & \multicolumn{2}{c}{CelebA} \\ 
		\cmidrule{3-8}
		                   &Weight                       & FID          & KID          & FID          & KID   &  FID          & KID         \\ 
		\midrule
		WGAN               & –                           & 44.11 (1.16) & 25 (1)       & 38.61 (0.43) & 23 (1)  & 17.85 (0.69) & 12 (1)      \\
		WGAN-GP            & –                           & 35.91 (0.30) & 19 (1)       & 27.85 (0.81) & 15 (1)  & 10.03 (0.37) & 6 (1)    \\
		MMD-GAN            & 5-RBF                       & 41.28 (0.54) & 23 (1)       & 35.76 (0.54) & 21 (1)  & 18.48 (1.60) & 12 (1)    \\
		MMD-GAN-GP-L2      & 5-RQ                        & 38.88 (1.35) & 21 (1)       & 31.67 (0.94) & 17 (1)  & 13.22 (1.30) & 8 (1) \\ 
		\midrule
		\multirow{2}{*}{CF-GAN}
		& $\mathcal{N}_{(\sigma=0.5)}$    & 39.81 (0.93) & 23 (1)       & 33.54 (1.11) & 19 (1)  & 13.71 (0.50) & 9 (1)    \\
		& $\mathcal{T}_{(\sigma=1)}$      & 41.41 (0.64) & 22 (1)       & 35.64 (0.44) & 20 (1)  & 16.92 (1.29) & 11 (1)    \\ 
		\cmidrule{2-8}
		\multirow{4}{*}{OCF-GAN}
		& $\mathcal{N}_{(\hat{\sigma})}$  & 38.47 (1.00) & 20 (1)       & 32.51 (0.87) & 19 (1)  & 14.91 (0.83) & 9 (1)    \\
		& $\mathcal{T}_{(\hat{\sigma})}$  & 37.96 (0.74) & 20 (1)       & 31.03 (0.82) & 17 (1)  & 13.73 (0.56) & 8 (1)    \\
		& $\mathcal{L}_{(\hat{\sigma})}$  & 36.90 & 20       & 32.09  & 18  & 14.96 & 10    \\
		& $\mathcal{U}_{(\hat{\sigma})}$  & 37.79 & 21       & 31.80 & 18  & 14.94 & 10    \\
		\cmidrule{2-8}
		\multirow{1}{*}{CF-GAN-Smooth} & $\mathcal{N}_{(\sigma=0.5)}$ &  41.17  &  24  & 32.98 &  19 & 13.42 & 9   \\
		OCF-GAN-Smooth & $\mathcal{N}_{(\sigma)}$ &  38.97 & 21   & 32.60 & 18  &  14.97 &  9   \\
		\cmidrule{2-8}
		\multirow{4}{*}{OCF-GAN-GP}
		& $\mathcal{N}_{(\hat{\sigma})}$  & \textbf{33.08 (0.26)}  &  \textbf{17 (1)}  & \textbf{26.16 (0.64)} &   \textbf{14 (1)}  & \textbf{9.39 (0.25)}  & \textbf{5 (1)}  \\
		& $\mathcal{T}_{(\hat{\sigma})}$  & 34.33 (0.77)  &  18 (1)  & 26.86 (0.38) &   15 (1)  & 9.61 (0.39)  &  6 (1)   \\
		& $\mathcal{L}_{(\hat{\sigma})}$  & 36.06  &  19  & 29.31 &   16  & 11.65  &  7    \\
		& $\mathcal{U}_{(\hat{\sigma})}$  & 35.14  &  18  &  27.62 &   15 & 10.29   &  6   \\
		\bottomrule
	\end{tabular}
\end{table}

\begin{table}
	\centering
	\small
	\caption{FID and KID scores (lower is better) achieved by the various models for the MNIST dataset. Results are averaged over 5 random runs and the standard deviation is indicated in parentheses.}
	\label{tab:mnist}
	\begin{tabular}{llrr} 
		\toprule
		\multirow{2}{*}{Model}                        & \multirow{2}{*}{Kernel/Weight}  & \multicolumn{2}{c}{MNIST}  \\ 
		\cmidrule{3-4}
		&                                 & FID         & KID $\times 10^3$         \\ 
		\midrule
		WGAN                                          & –                               & 1.69 (0.09) & 20 (2)       \\
		WGAN-GP                                       & –                               & 0.26 (0.02) & 2 (1)       \\
		MMD-GAN                                       & 5-RBF                       & 0.68 (0.18) & 10 (5)       \\
		MMD-GAN-GP$_{L2}$                              & 5-RQ                        & 0.51 (0.04) & 6 (2)       \\ 
		\midrule
		\multirow{2}{*}{CF-GAN} & $\mathcal{N}_{(\sigma=1)}$      & 0.98 (0.33) & 16 (10)        \\
		& $\mathcal{T}_{(\sigma=0.5)}$    & 0.85 (0.19) & 12 (4)       \\ 
		
		\cmidrule{2-4}
		\multirow{2}{*}{OCF-GAN}
		& $\mathcal{N}_{(\hat{\sigma})}$  & 0.60 (0.12) & 7 (3)        \\
		& $\mathcal{T}_{(\hat{\sigma})}$  & 0.78 (0.11) & 9 (1)       \\
				\cmidrule{2-4}
		\multirow{2}{*}{OCF-GAN-GP}
		& $\mathcal{N}_{(\hat{\sigma})}$  & 0.35 (0.02) & 3 (1)        \\
		& $\mathcal{T}_{(\hat{\sigma})}$  & 0.48 (0.06) & 6 (1)       \\
		\bottomrule
	\end{tabular}
\end{table}

%\begin{figure*}
%	\centering
%	\subfloat[WGAN]{\includegraphics[width=0.4\linewidth]{mnist-wgan}}\quad
%	\subfloat[WGAN-GP]{\includegraphics[width=0.4\linewidth]{mnist-wgangp}}\quad\\
%	\subfloat[MMD-GAN]{\includegraphics[width=0.4\linewidth]{mnist-mmdgan}}\quad
%	\subfloat[MMD-GAN-GP]{\includegraphics[width=0.4\linewidth]{mnist-mmdgangp}}\quad\\
%	\subfloat[OCF-GAN-GP]{\includegraphics[width=0.4\linewidth]{mnist-ocfgangp}}\quad
%	\subfloat[MNIST Test Set]{\includegraphics[width=0.4\linewidth]{mnist-test}}\\
%	\caption{Random samples from the different models for MNIST dataset.}
%\end{figure*}


\end{document}